\pdfoutput=1

\documentclass[11pt]{article}

\usepackage{acl}

\usepackage{times}
\usepackage{latexsym}

\usepackage[T1]{fontenc}

\usepackage[utf8]{inputenc}

\usepackage{microtype}

\usepackage{inconsolata}
\usepackage{booktabs}
\usepackage{balance}
\usepackage{graphicx}
\usepackage{subcaption}
\usepackage{amsthm}

\usepackage{xcolor}
\usepackage{multirow}
\usepackage{array}
\usepackage{breqn}
\usepackage{mathtools}
\usepackage{listings}
\usepackage{amsmath}
\usepackage[ruled,vlined,linesnumbered]{algorithm2e}
\usepackage{tikz}
\usepackage[framemethod=TikZ]{mdframed}
\title{Penetrative AI: Making LLMs Comprehend the Physical World}

\author{Huatao Xu$^{1}$, Liying Han$^{2}$, Qirui Yang$^{1}$, Mo Li$^{1}$, Mani Srivastava$^{2}$\thanks{\ \ Mo Li and Mani Srivastava are the corresponding authors. Huatao Xu was a Ph.D. student at Nanyang Technological University.}\\
  $^{1}$Hong Kong University of Science and Technology, $^{2}$University of California Los Angeles \\
  \texttt{huatao@ust.hk, \{liying98, mbs\}@ucla.edu, \{qyangau,lim\}@cse.ust.hk}
  }
\begin{document}
\maketitle
\begin{abstract}
Recent developments in Large Language Models (LLMs) have demonstrated their remarkable capabilities across a range of tasks. Questions, however, persist about the nature of LLMs and their potential to integrate common-sense human knowledge when performing tasks involving information about the real physical world. This paper delves into these questions by exploring how LLMs can be extended to interact with and reason about the physical world through IoT sensors and actuators, a concept that we term "\textit{Penetrative AI}\footnote{Project website: \url{https://hkustwands.github.io/penetrative-ai/}}". The paper explores such an extension at two levels of LLMs' ability to penetrate into the physical world via the processing of sensory signals. Our preliminary findings indicate that LLMs, with ChatGPT being the representative example in our exploration, have considerable and unique proficiency in employing the embedded world knowledge for interpreting IoT sensor data and reasoning over them about tasks in the physical realm. Not only this opens up new applications for LLMs beyond traditional text-based tasks, but also enables new ways of incorporating human knowledge in cyber-physical systems.
\end{abstract}

\section{Introduction}
Large Language Models (LLMs) have made remarkable strides \citep{brown2020language, scao2022bloom, zeng2022glm}. A particularly revolutionary milestone is ChatGPT \citep{openai2023gpt4}, which excels in fluid, human-like conversations, marking a new era in human-AI interactions. These latest LLMs cultivated on extensive text datasets have showcased remarkable capabilities across diverse tasks, including coding and logical problem-solving \citep{creswell2022selection}. These out-of-the-box capabilities have demonstrated that they already comprise enormous amounts of world knowledge \footnote{Some studies referred to it as a world model \citep{lecun2022path} of how the world works.}.
\begin{figure}[t!]
  \centering
  \includegraphics[width=0.9\linewidth]{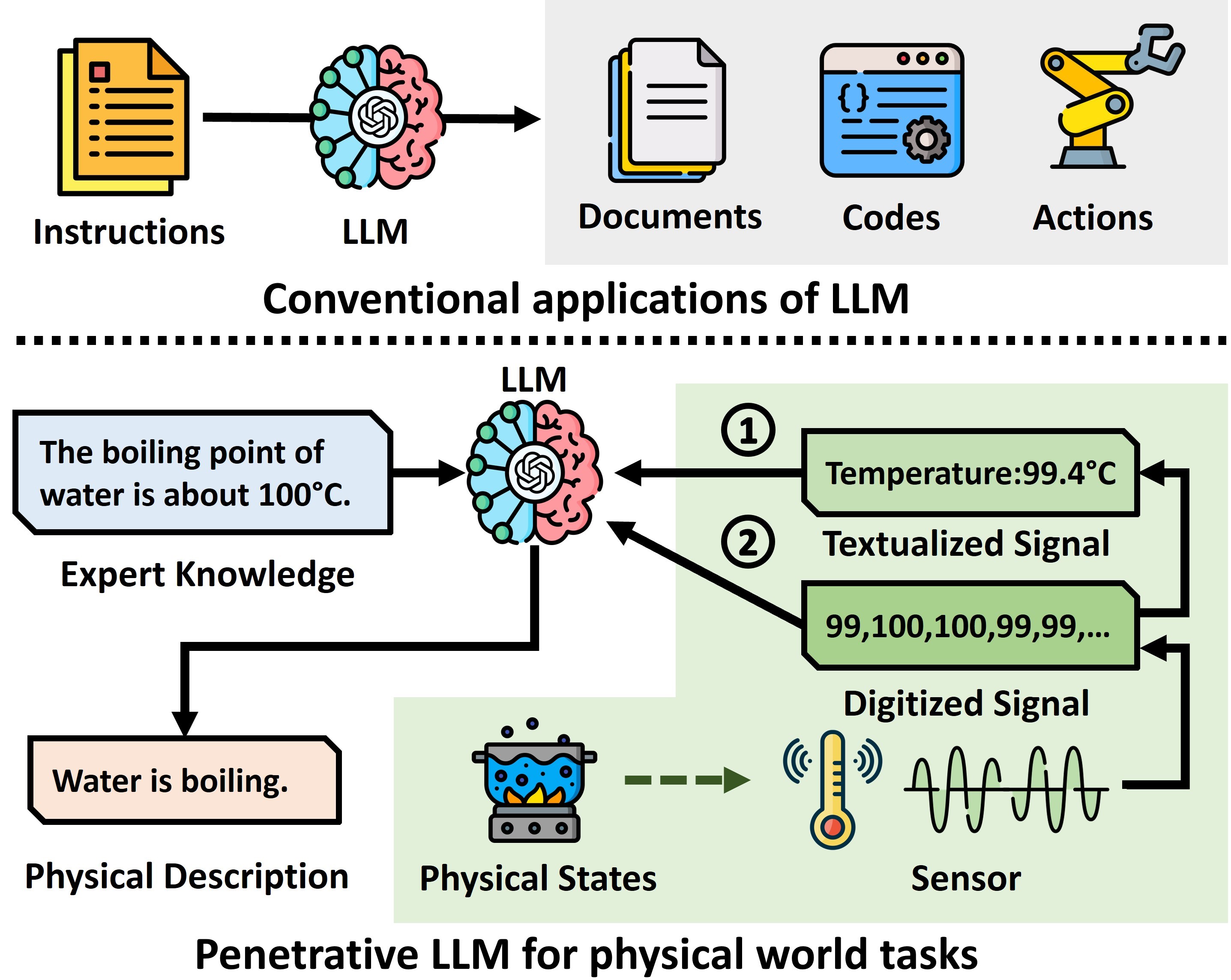}
  \caption{Overview of Penetrative AI.}
  \label{fig:objective}
\end{figure}

This paper is motivated by an essential and intriguing question: can we enable LLMs to complete tasks in the real physical world? We delve into this inquiry and explore extending the boundaries of LLMs' capabilities by directly letting them interact with the physical world through Internet of Things (IoT) sensors. A basic example of this process is depicted in Figure \ref{fig:objective}, where different from the conventional way of LLMs, an LLM is expected to analyze sensor data which are indeed projections from the physical world. We conjecture that LLMs, having been trained on vast amounts of human knowledge, learned the physical world which can be directly harnessed for analysis of such sensory information to derive deep insights that traditionally require background knowledge from human experts and/or bespoke machine learning models trained with large amounts of labeled sensor data. 
\begin{figure*}[t!]
  \centering
  \includegraphics[width=1.0\linewidth]{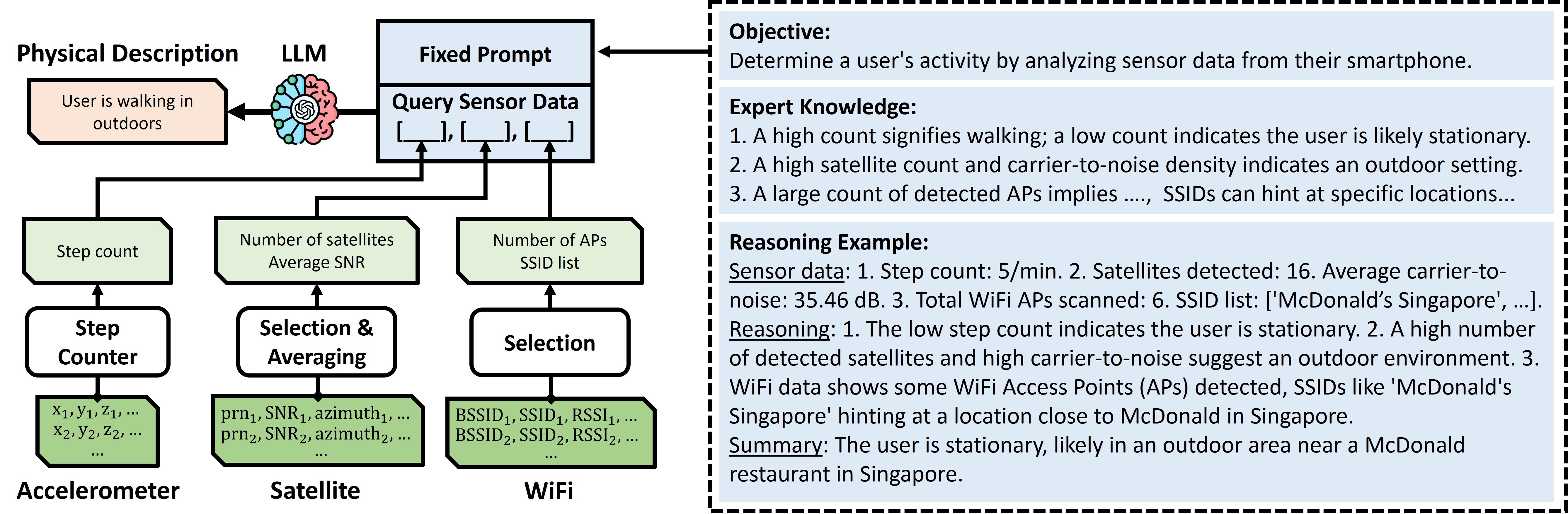}
  \caption{Overview of user activity sensing with LLMs.}
  \label{fig:as}
\end{figure*}

As illustrated in Figure \ref{fig:objective}, we formulate such a problem from a signal processing's point of view, and specifically explore the LLMs' penetration into the physical world at two signal processing levels with the sensor data: i) with the textualized signals derived from underlying sensor data, and ii) with the digitized signals, essentially numerical sequences of raw sensor readings. We term this endeavor "\textbf{\textit{Penetrative AI}}" -- where the embedded world knowledge in LLMs serves as a foundation model, seamlessly integrated with the Cyber-Physical Systems (CPS) for perceiving and intervening in the physical world.

Our methodology is exemplified through two illustrative applications at two different levels, respectively - user activity sensing where textualized signals from smartphone accelerometer, satellite, and WiFi data are analyzed to discern user motion and environment conditions, and human heartbeat detection where digitized electrocardiogram (ECG) data are utilized to derive the heartbeat rate. Preliminary findings are encouraging, showcasing LLMs' proficiency in interpreting IoT sensor data and performing perception tasks in the physical world. Our exploration also underscores that existing LLMs like ChatGPT-4 may already possess the capability to establish intricate connections among world knowledge and can be guided to tackle CPS tasks.

Section 2 and Section 3 will elaborate on the design and experiment results of these two illustrative applications. Section 4 sets the scope of penetrative AI and shares our insights on the foreseeable challenges to advance this burgeoning research frontier. We present related works in Section 5 and conclude this paper in Section 6.

\section{Penetrative LLM with Textualized Signals}
This section describes tasking LLMs to comprehend sensor data at the textualized signal level.

\subsection{An Illustrative Example}
We take activity sensing as an illustrative example, where LLMs interpret sensor data collected from smartphones to derive user activities. The input sensor data encompass smartphone accelerometer, satellite, and WiFi signals, and the desired output is to discern the user motion and environment context. Figure \ref{fig:as} presents the overview of this LLM-based design -- the sensor data are pre-processed by individual sensing components and the textualized sensor states are supplied to the LLM with a fixed prompt for activity inference.

\textbf{Objective and Rationale}. 
We convey a clear task to LLMs -- "determine a user's motion and surrounding conditions by analyzing sensor data from their smartphones". The basic idea is that when the user conducts different activities in different environments, the collected sensor data would exhibit varied patterns, which reveal the users' activities. 

\textbf{Data Preparation}. 
To facilitate LLMs comprehension of the sensor data, we undertake a preprocessing step where raw data from different sensing modules are separately converted into textualized states that are expected interpretable by LLMs. Figure \ref{fig:as} illustrates such a step. 

\begin{figure*}[t!]
  \centering
  \includegraphics[width=1.0\linewidth]{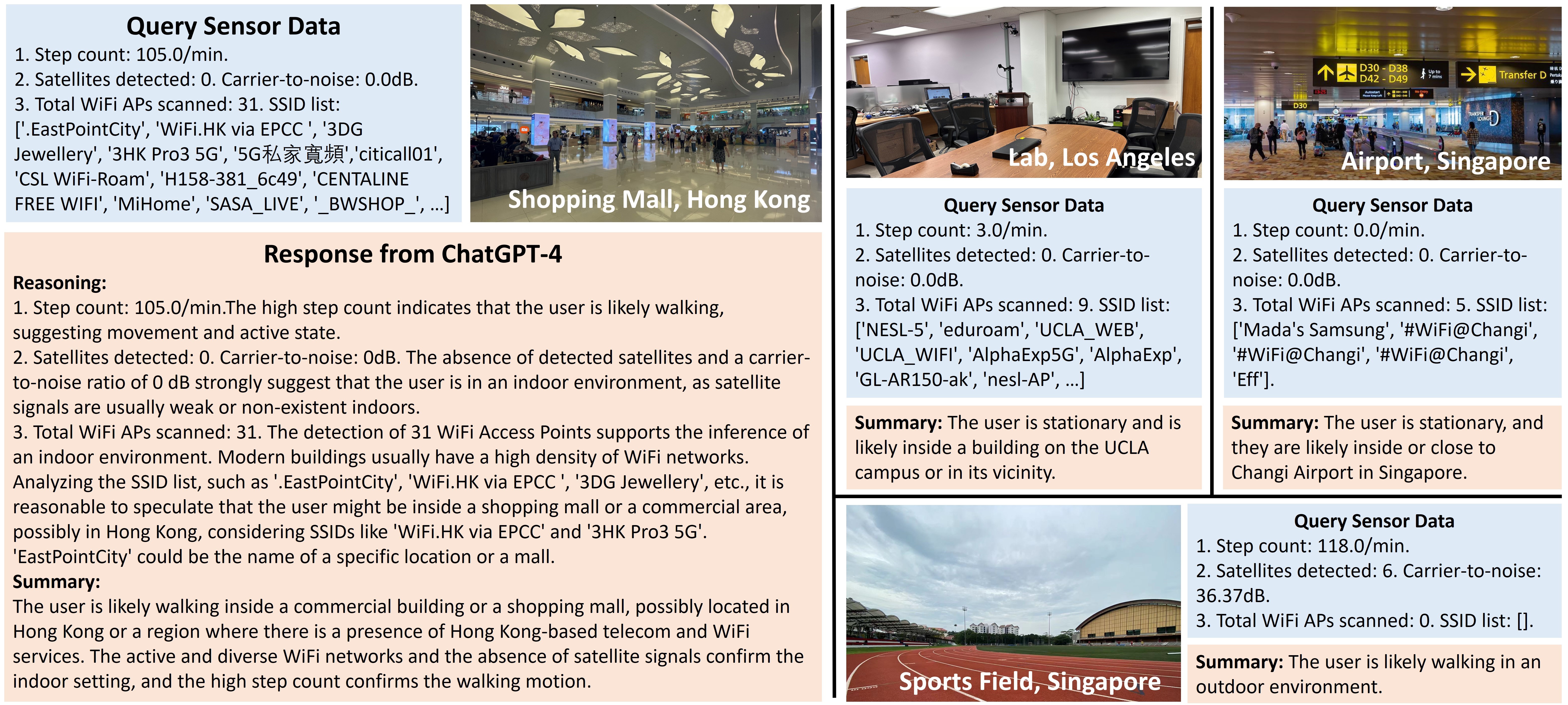}
  \caption[]{Response examples of ChatGPT-4 for activity sensing.}
  \label{fig:as:exp}
\end{figure*}

To pre-process long accelerometer readings (6,000 samples from 10 seconds of triaxial accelerations sampled at 200 Hz), we employ the Android step detector, which is a built-in step-counting implementation \citep{android@motion} and can transform the 6,000 raw data points into a single textually expressed state, e.g., "step count: 5/min".

The Android system also offers a comprehensive set of Global Navigation Satellite System (GNSS) satellite measurements \citep{android@gnss}, including information like Pseudo-Random Noise as a satellite identifier, Signal-to-Noise Ratio (SNR), and many others. To streamline the data for LLMs interpretation, we filter and distill the satellite data into two key attributes: the number of detected satellites and their average SNR.

The Android system supports scanning for nearby APs and provides comprehensive information about scanned APs \citep{android@wifi}. Similar to satellite data, we disregard less relevant details and focus on critical information -- Service Set Identifier (SSID) and Received Signal Strength Indicator (RSSI). To streamline the data and reduce text length, we further filter APs with an RSSI lower than -70 and instruct LLMs to analyze the SSIDs to capture useful location information.

\textbf{Expert Knowledge}. 
We guide LLMs by including explicit text-based descriptions of the relationship between sensor patterns and user activity states in the prompts, as illustrated in Figure \ref{fig:as}. For instance, a high satellite count and carrier-to-noise density indicate an outdoor setting with strong satellite signals.

\textbf{Reasoning Examples}.
Following expert knowledge, we can provide reasoning examples to enhance the proficiency of LLMs. Each example includes the data for processing, a step-by-step reasoning process, and a brief summary of the ground truth context, which adopts the chain of thought (CoT) \citep{wei2022chain} prompting. Figure \ref{fig:as} illustrates this with the reasoning example section.

\textbf{Complete Prompt}.
A full prompt includes a defined objective and expert knowledge of the sensor data, all in natural language as demonstrated in Figure \ref{fig:as}. Essentially, the way we construct the prompt serves as a means to educate and instruct LLMs to interpret sensor data and formulate its answers into a concise format. We thereafter present the prompt with succinct textualized sensor data of novel queries to LLMs as shown in  Figure \ref{fig:as}, which we expect to generate the inference results as a concise description of the user's activity. Note that the prompt, once completed, is frozen and we simply supply new textualized sensor data for new inferences without altering the prompt any further.

\subsection{Experiment Results}
\begin{table*}[ht!]
\centering
\caption{Overall performance of LLMs in activity sensing. 'e.k.' indicates the expert knowledge and 'exam.' indicates a reasoning example.}
\scalebox{0.95}{
\begin{tabular}{l|l|ccc|ccc}
\toprule
\multicolumn{1}{c|}{\multirow{2}{*}{Task}}    & \multicolumn{1}{c|}{Metric}   & \multicolumn{3}{c|}{Failure Rate ($\downarrow$)} & \multicolumn{3}{c}{Classification Accuracy ($\uparrow$)} \\ \cmidrule{2-8} 
& \multicolumn{1}{c|}{Prompt} & plain   & w/ e.k.  & w/ e.k. +1 exam.  & plain     & w/ e.k.     & w/ e.k. +1 exam.    \\ \midrule 
\multicolumn{1}{c|}{\multirow{3}{*}{\begin{tabular}[c|]{@{}c@{}}Motion\\ Detection\end{tabular}}} & PaLM 2    &   0\%     &    0\%      &   0\%               &   1.00        &      1.00      &           1.00         \\ 
& ChatGPT-3.5 &    3\%     &   0\%       &        0\%          &           0.97     &   1.00            &     1.00   \\ 
& ChatGPT-4   &    0\%   &    0\%      &       0\%           &           1.00     &   1.00            &   1.00     \\ \midrule 
\multicolumn{1}{c|}{\multirow{3}{*}{\begin{tabular}[c|]{@{}c@{}}Indoor/outdoor\\ Detection\end{tabular}}} & PaLM 2    &   0\%     &   0\%      &  0\%   &  0.79    &    0.88    &     0.91      \\ 
& ChatGPT-3.5 &     15\%     &  0\%   &     3\%      &    0.70     &   0.82     &  0.88    \\ 
& ChatGPT-4   &    0\%     &    0\%      &         0\%         &           0.88     &   0.91             &    \textbf{0.94}    \\
\bottomrule
\end{tabular}
}
\label{tab:exp:as}
\end{table*}
We conduct experiments in various scenarios -- on university campuses, commercial buildings, subway stations, outdoor spaces, and across different cities. The dataset contains 165 instances and all sensor data are collected using a Samsung Galaxy S8 Android smartphone. Accelerometer data are sampled at 100 Hz, while the satellite and WiFi data are sampled at 0.2 Hz. Sensor data are gathered from time windows spanning durations of 10 to 60 seconds and the latest satellite and WiFi scanning results are adopted. The evaluation is carried out using PaLM 2 \citep{anil2023palm}, ChatGPT-3.5 (\texttt{gpt-3.5-turbo-0613}) and ChatGPT-4 (\texttt{gpt-4-0613}) \citep{openai2023gpt4}, accessible through the official API with default parameter settings. 


Figure \ref{fig:as:exp} shows several example answers of ChatGPT-4 together with ground-truth contexts. Due to space limits, we only show the detailed response for the first case. The results suggest ChatGPT-4's ability to identify user motion and indoor/outdoor states with the provided textualized sensor data. Additionally, it demonstrates an impressive capacity to deduce intricate details about the user's surroundings, e.g., it reasons that the user is likely inside a shopping mall by analyzing the scanned WiFi SSIDs in the first case.

To quantitatively assess the efficacy of such an approach, we tasked LLMs to explicitly provide the states of motion (between "stationary" and "motion") and environment (between "indoors" and "outdoors"), which are identified by a keyword detection process. We experiment with varied settings -- plain, with additional expert knowledge, as well as with the additional reasoning example in the prompt. To assess the performance of the penetrative LLMs, we utilize two key metrics: the failure rate and classification accuracy. In our cases, "failure" refers to instances where the LLMs are unable to generate valid states relevant to the task. The failure rate is thus calculated as the proportion of such instances to the total number of cases.
\begin{figure*}[t!]
  \centering
  \includegraphics[width=1.0\linewidth]{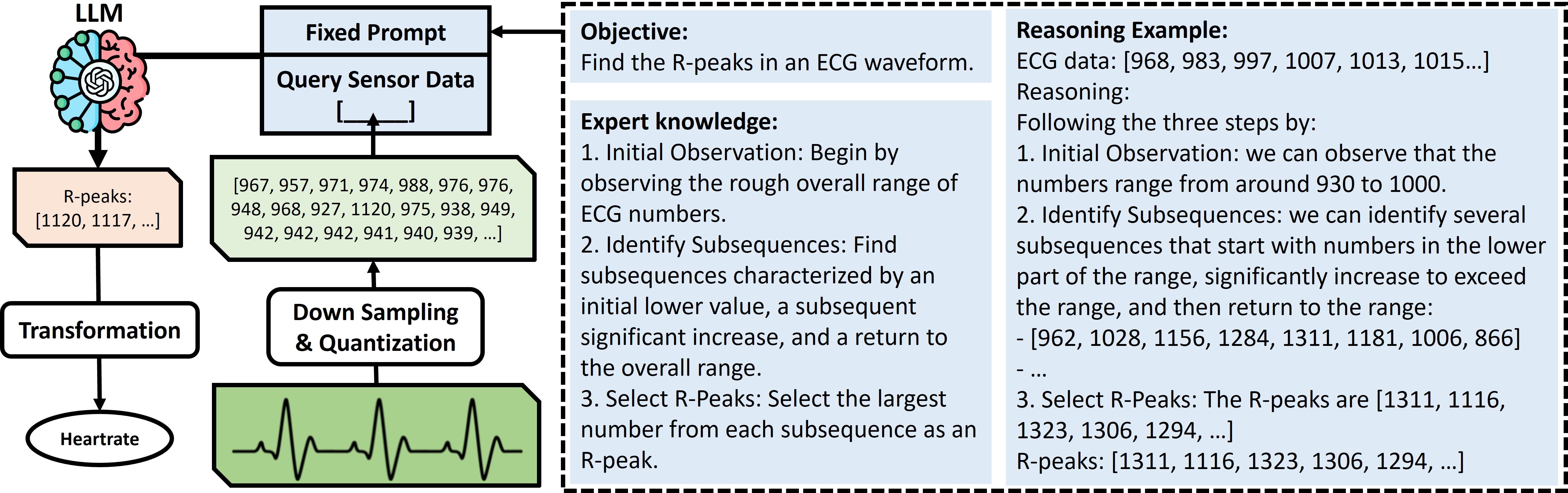}
  \caption{Overview of heart rate detection with LLMs.}
  \label{fig:hd}
\end{figure*}

Table \ref{tab:exp:as} summarizes the overall performance of different LLMs on the two tasks. ChatGPT-3.5 occasionally outputs 'unknown' states leading to higher failure rates in the two tasks. This rate can be effectively reduced to 0\% by incorporating expert knowledge. The results show three models perform reasonably well in the motion detection task. The task of discerning indoor/outdoor is more challenging, largely due to its reliance on the fusion of multimodal sensor data. Nevertheless, a notable enhancement is achieved when prompts are enriched with expert knowledge and one reasoning example. PaLM 2 and ChatGPT-4 achieve above 90\% accuracy with the best prompt scheme.

The location context detection using WiFi SSIDs demands a broad base of general knowledge. In many cases, WiFi signals are either sparse or the SSIDs lack unique identifiers that could be leveraged to infer detailed location contexts. To provide the quantitative evaluation for this task, we manually assess all instances and identify that among the 143 instances with scanned WiFi signals, 115 feature SSIDs informative enough to deduce location contexts, whereas 22 do not. Table \ref{tab:exp:ha:loc} presents the precision and recall for three models in the location context detection task. To further evaluate the quality of the location contexts generated by LLMs, we provide a reference location context based on the actual environments and compare sentence-level similarities between this reference and the generated contexts using chrF \citep{popovic2015chrf} and BERTScore \citep{zhang2019bertscore} F1 score. The results shown in Table \ref{tab:exp:ha:loc} indicate ChatGPT-4 achieves the best overall performance with good chrF and BERTScore.
\begin{table}[t!]
\centering
\caption{Performance of LLMs on location context detection using WiFi SSIDs.}
\scalebox{0.8}{
\begin{tabular}{l|ccc}
\toprule
Metric    & PaLM 2 & ChatGPT-3.5 & ChatGPT-4 \\\midrule
Precision & 0.93   & \textbf{1.00}        & \textbf{1.00}          \\
Recall    & 0.96   & 0.48        & \textbf{0.97}       \\ \midrule
chrF &  0.39   &   0.36      & \textbf{0.59} \\
BERTScore & 0.65   & 0.58        & \textbf{0.71} \\
\bottomrule
\end{tabular}
}
\label{tab:exp:ha:loc}
\end{table}

Overall, the above experiment results suggest LLMs can be effective in analyzing sensor signals when properly abstracted into textual representations.

\section{Penetrative LLM with Digitized Signals}
This section describes our effort that goes beyond the general expectations of the textualized signal processing ability of LLMs. We specifically study the potential of LLMs in comprehending digitized sensor signals.

\subsection{An Illustrative Example}
We take human heart rate detection as an illustrative example, where we task LLMs with the input of ECG waveforms to identify the R-peaks, based on which we can then derive the heartbeat rate. Fundamentally different from the previous example, all sensor data in this application are expressed as sequences of digitized samples. Figure \ref{fig:hd} provides an overview of the design.

\textbf{Objective and Rationale}.
The sensor data consist of a numerical sequence representing an ECG waveform.  Our objective for LLMs is to identify the "R-peaks" \citep{yanowitz2010lesson}, which are tall upward deflections in ECG data and correspond to the red dots in Figure \ref{fig:hd}. The objective part of the prompt succinctly states: "Find the R-peaks in an ECG waveform". An interesting and challenging job in this application is, we incorporate expert knowledge directly into the prompts, delegating the signal processing task to LLMs.

\textbf{Data Preparation}.
The original ECG data are collected at a high sampling rate, e.g., 360Hz. In our design, raw ECG readings are down-sampled to 72 Hz and quantized to their integer parts to reduce the length and complexity of the sequence. 

\textbf{Expert Knowledge}.
To assist the LLMs, we try to give a detailed description of R-peaks with the context of QRS complex \citep{kadambe1999wavelet} in the prompt, i.e., "The QRS complex, a recurring feature in ECG data, signifies the ventricles' consistent depolarization in the heart. It comprises the Q, R, and S waves, where the Q wave shows a downward deflection, followed by an upward-moving R wave, and then the S wave, which deflects downward after the R wave. The maximum amplitude of the R wave is known as the R-peak.". 

Our experiments show that it remains challenging for LLMs to perform the task for a long sequence of ECG digits with only descriptions of R-peaks. By observing the patterns of R-peaks, we instead design a procedure that LLMs understand to guide the selection of R-peaks. As depicted in Figure \ref{fig:hd}, three steps are included: 1) assessing the overall range of ECG numbers, 2) identifying subsequences characterized by an initial lower value, a subsequent significant increase, and a return to the overall range, and 3) selecting the highest value from each such subsequence as the R-peak. We examine whether LLMs like ChatGPT can effectively execute such a fuzzy logic (without explicit thresholding) when processing the digitized signals.

\textbf{Reasoning Examples}.
We also furnish LLMs with illustrative examples as shown in Figure \ref{fig:hd}, which encompasses the digitized ECG data, a reasoning procedure, and a summary of R-peaks (check more details in Appendix \ref{sec:app:pro}). 


\begin{figure}[t!]
  \centering
  \includegraphics[width=1.0\linewidth]{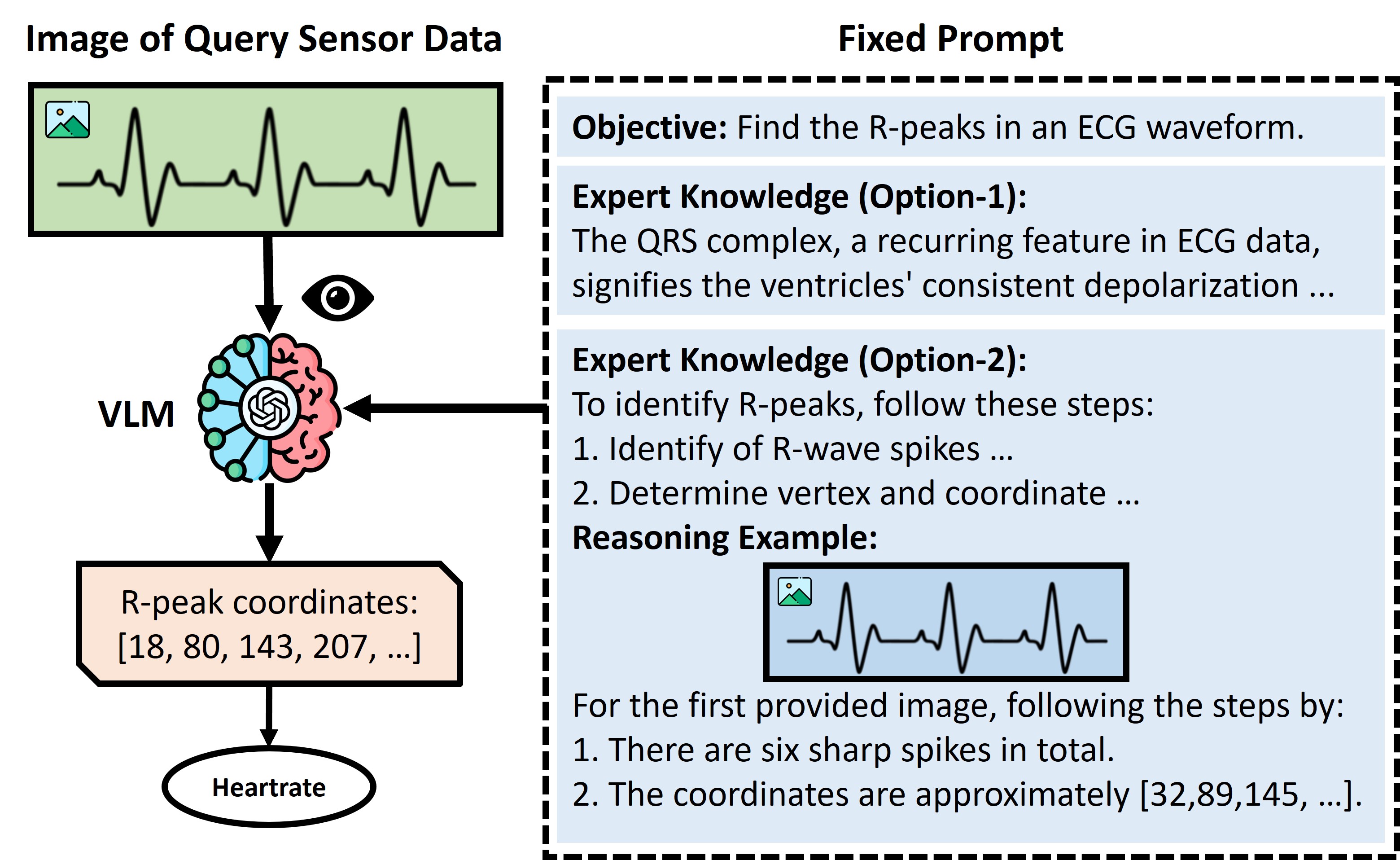}
  \caption{Overview of heart rate detection with VLMs.}
  \label{fig:had:img}
\end{figure}


\begin{table*}[ht!]
\centering
\caption{Performance comparison in heart rate detection. The upper part shows the MAE ($\downarrow$) of conventional signal processing methods while the lower part includes the hallucination rates ($\downarrow$) and MAEs ($\downarrow$) of penetrative LLMs/VLMs. "description" means the description of R-peaks, "proc." indicates the inclusion of detailed processing procedure, and "exam." indicates the inclusion of reasoning examples. }
\scalebox{0.90}{
\begin{tabular}{cccccc}
\toprule
\multicolumn{1}{l|}{Window Size} &  Pan–Tompkins & Hamilton       & Christov & TMA                & SWT            \\ \midrule 
\multicolumn{1}{c|}{5 seconds}                 & 5.76         & \textbf{3.60}           & 7.08     & 9.24                & 4.20           \\ \midrule %
\multicolumn{1}{c|}{30 seconds}                 & 1.06         & 0.76           & 1.30     & 1.64      &   \textbf{0.37}    \\ \midrule\midrule 
\multirow{2}{*}{LLM/VLM}                     & \multicolumn{5}{c}{Prompt Scheme (5-second window size)}        \\\cmidrule{2-6}
               & w/ description & w/ proc.   & w/ proc. + 1 exam. & w/ proc. + 2 exam. & one-shot \\ \midrule 
\multicolumn{1}{l|}{PaLM 2}        &       95\%, 816.00        &95\%, 148.80     &       58\%, 30.29        & 50\%, 82.32 & 97\%, 84.00 \\
\multicolumn{1}{l|}{ChatGPT-3.5} &      22\%, 329.92   & 14\%, 187.95     &         10\%, 64.27   & 2\%, 20.96   & 27\%, 579.12  \\  
\multicolumn{1}{l|}{ChatGPT-4}  &    0\%, 81.84      &    0\%, 92.40      &       \textbf{0\%, 1.56}   & 0\%, 4.80  & 0\%, 142.68 \\ \midrule
\multicolumn{1}{l|}{GPT-4V}  &    0\%, 9.60      &    0\%, 12.61      &      0\%, 8.16   & 0\%, 11.16 & 0\%, 12.48\\ 
\bottomrule
\end{tabular}
}
\label{tab:exp:hd}
\end{table*}
\subsection{Digitized Data as Figures}

Following the concept of Penetrative AI, we further test using Vision-Language Models (VLMs) \citep{radford2021learning, jia2021scaling,lu2019vilbert,tan2019lxmert}, which are vision interfaced LLMs, to "see" digitized sensor data as figures and accordingly execute real-world tasks. Figure \ref{fig:had:img} illustrates the design with VLM to process the same R-peak detection example.

In this exploration, ECG data are visualized in the figures and fed to VLMs, which are tasked with locating the coordinates of R-peaks in such figures. Figure \ref{fig:had:img} illustrates the process. The objective is to count the R-peaks in the ECG data and only a general description of R-peaks is provided as expert knowledge. Different prompt schemes are also tested where a more detailed procedure to detect R-peaks with one or more reasoning examples containing reference ECG figures (see Appendix \ref{sec:app:pro} for detailed illustration). We investigate the efficacy of VLMs, GPT-4V \citep{gpt4_system_card} in this study, in performing perceptual tasks.

\subsection{Experiment Results}


We conduct experiments with the MIT-BIH Arrhythmia Database \citep{goldberger2000physiobank}, which is an ECG dataset with ground truth annotations for R-peaks. We downsampled the raw ECG signal to 72 Hz and each ECG query is from a 5-second window comprising 360 numerical values by default. The evaluation is carried out using the three models, i.e., PaLM 2, ChatGPT-3.5, and ChatGPT-4 with default parameters. The experiment is also performed with digitized ECG figures using GPT-4V \citep{gpt4_system_card} (\texttt{gpt-4-vision-preview}). For comparison, we also test the performance of classical signal processing approaches \citep{bernd_porr_2023_7652725}, including Pan-Tompkins \citep{pan1985real}, Hamilton \citep{hamilton2002open}, Christov \citep{christov2004real}, Two Moving Average (abbreviated as TMA) \citep{elgendi2010frequency}, and Stationary Wavelet Transform (abbreviated as SWT) \citep{kalidas2017real}. We use the Mean Absolute Error (MAE) to measure the error in beats per minute between the detected and actual heart rates.

The experiments with LLMs/VLMs are conducted with different prompt schemes, including (i) containing only general descriptions of R-peaks, (ii) containing a detailed detection procedure, (iii) containing the procedure as well as varied numbers of reasoning examples, and (iv) one-shot prompting \citep{liu2023large} containing the one example of ECG data and actual R-peak values. In our evaluation, "hallucination" is defined as cases where the LLMs/VLMs are unable to proceed or generate R-peak outputs. The MAE is averaged across cases where models can produce R-peak outputs.

\textbf{Overall Performance.} Table \ref{tab:exp:hd} summarizes the performance of various baseline methods alongside four penetrative LLM/VLMs in the task. We observe that conventional signal processing baselines give high MAEs when the window size of query data is 5 seconds, which can be significantly improved when the window size increases to 30 seconds. However, the performance of LLMs varies a lot. PaLM 2, for instance, frequently repeats the query sensor data in the response, leading to high hallucination rates and significant MAEs. ChatGPT-3.5 shows a reduction in hallucination rates but tends to produce extended sequences of R-peaks, resulting in significant errors. 

Remarkably, ChatGPT-4 completely avoids hallucinations and yields an impressive MAE of 1.56 when the prompt is incorporated with a dedicated procedure and one reasoning example. This performance is noteworthy, as it surpasses all conventional signal processing baselines with 5-second data. It is noteworthy, however, that ChatGPT-4 does not record the lowest MAE when provided with two reasoning examples. This phenomenon may be attributed to the increased task complexity due to additional examples. One more reasoning example encompasses an ECG sequence and a reasoning procedure, which occupies about 2,200 tokens for ChatGPT. 

GPT-4V exhibits commendable efficacy and it outperforms all LLMs when only a general description of R-peaks is provided in the prompt, demonstrating its potential for general adoptions where its usage is completely independent of any signal processing knowledge. 

In conclusion, our findings indicate that LLMs exemplified by ChatGPT-4 can exhibit remarkable proficiency in analyzing physical digitized signals when provided with proper guidance.

\begin{figure}
\begin{subfigure}{.48\linewidth}
  \centering
  \includegraphics[width=\linewidth]{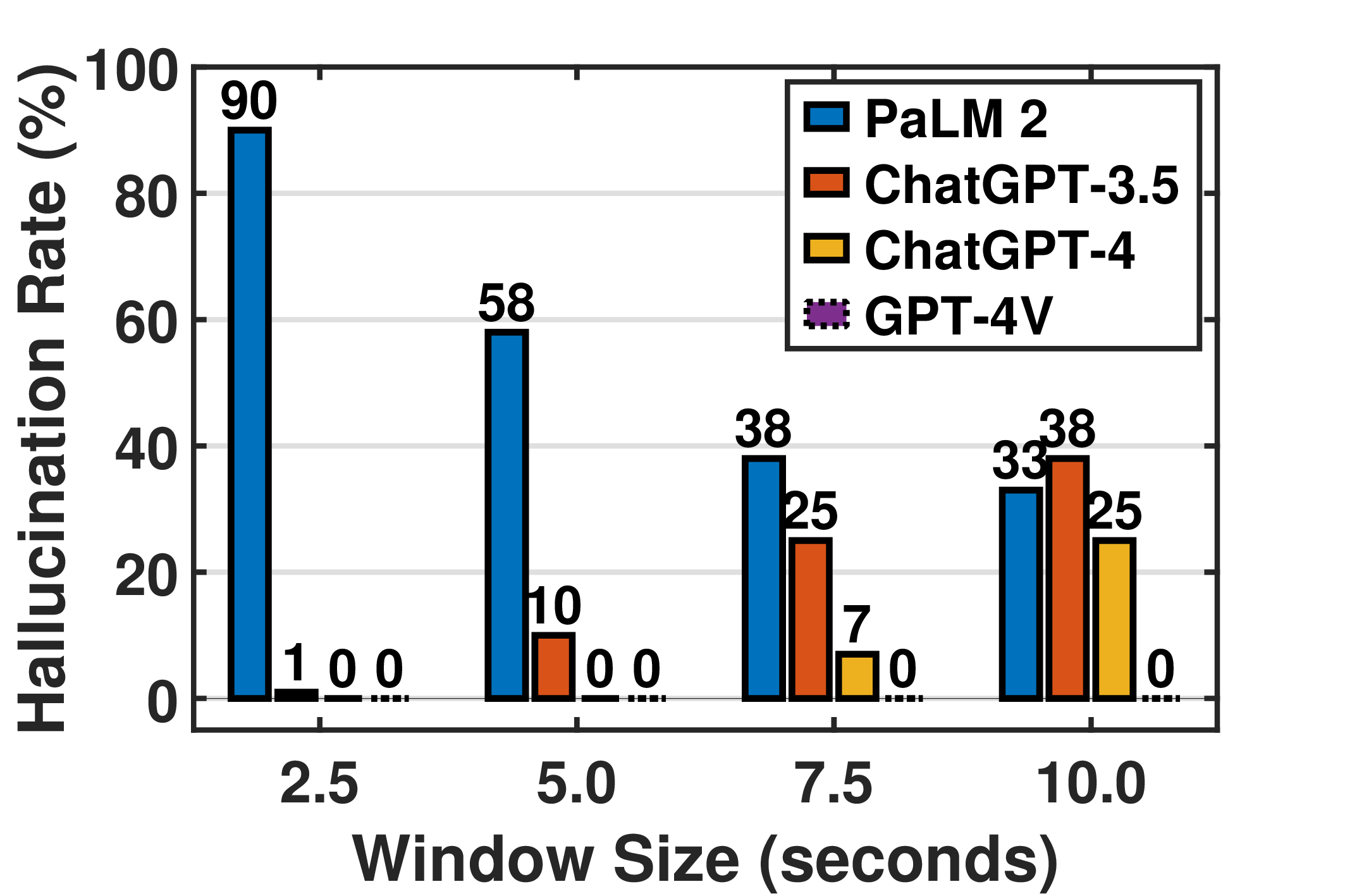}
  \caption{Hallucination rate.}
  \label{fig:exp:hd:1}
\end{subfigure}
\hspace{0.1em}
\begin{subfigure}{.48\linewidth}
  \centering
  \includegraphics[width=\linewidth]{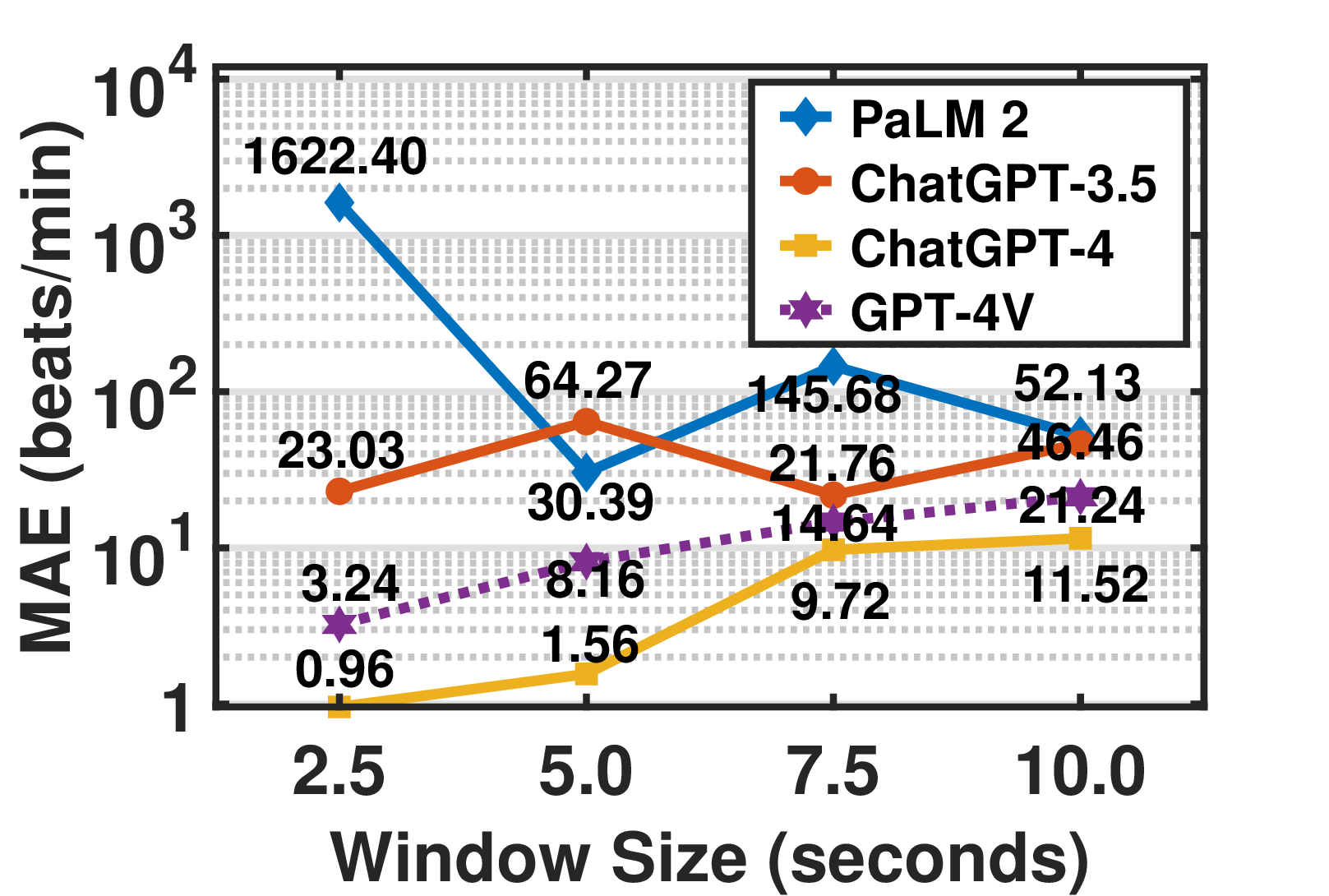}
  \caption{MAE.}
  \label{fig:exp:hd:2}
\end{subfigure}
\caption{Impact of window size of query sensor data.}
\label{fig:exp:hd}
\end{figure}
\textbf{Impact of Window Size.} We investigate how the window size of ECG query data impacts the end performance of LLM/VLMs. We adopt the prompt scheme encompassing the procedure and one reasoning example. As shown in Figure \ref{fig:exp:hd}, we vary the window size from 2.5 to 10 seconds, adjusting the reasoning example correspondingly for each window size. The stability of PaLM 2's performance is inconsistent, and a reduction in hallucination rates does not translate to lower MAEs. The hallucination rates for both ChatGPT-3.5 and ChatGPT-4 escalate with the increase in window size. A plausible explanation for such a trend is ChatGPT's inherent limitation in processing extensive lengths of digitized sequences. For instance, a 10-second window of query sensor data contains 720 numerical values, resulting in approximately 5,100 tokens. These findings suggest the inefficacy of existing LLMs like ChatGPT when tasked to process long digitized signals. The GPT-4V keeps a zero hallucination rate but exhibits a similar trend of MAE increase with bigger window sizes.




\section{Penetrative AI}
While not achieving perfect accuracy, LLMs exhibit surprisingly encouraging performance, even when dealing with pure digital signals. This presents an enticing opportunity to leverage LLMs' world knowledge as a foundation model to derive insights from sensory information while requiring no or little additional task knowledge or data, i.e., in zero or few-shot settings. Such a capability may be equipped with IoT sensors and actuators to build intelligence into cyber-physical systems -- a concept we term "\textit{Penetrative AI}".

\subsection{Scope}
"Penetrative AI" is concerned with exploring the foundation role of LLMs in completing tasks in the physical world. Two primary characteristics define its scope -- i) the involvement of the embedded world knowledge in LLMs \footnote{or variations like Vision-Language Models (VLMs) \citep{gpt4_system_card} which adapt to other input modalities.}, and ii) the integration with IoT sensors and/or actuators for perceiving and intervening the physical world. It is important to distinguish the scope of Penetrative AI from existing practices where the LLMs are not engaged with their world knowledge in direct analysis of sensor inputs or CPS control. Examples include classical NLP applications of LLMs, conventional machine learning adopted in CPS, and LLMs involved in the CPS loop but not applied to comprehending the physical world phenomena.

As the example applications demonstrate, penetrative AI may offer the following potentials. It simplifies solution deployment, allowing user-machine interaction in plain language and minimizing the need for extensive programming skills. It also enhances data efficiency as LLMs embedded with vast world knowledge can effectively generalize to new tasks. LLMs adeptly handle fuzzy logic well, drawing inferences from vague or disorganized information, and bypassing the need for precise logic. Finally, the penetrative AI offers an innovative opportunity for multimodal fusion, where diverse data types are transformed into a uniform text format, facilitating seamless adaptation to various tasks without extensive model re-engineering.

\subsection{Challenges and Future Directions}
Adopting LLMs in a penetrative way for CPS is non-trivial since LLMs are typically trained with extensive text corpora for NLP applications and thus may lack expertise and domain knowledge for CPS tasks. Unleashing its full potential necessitates addressing the challenges as follows:

\textbf{Understanding the knowledge boundaries of LLMs.} A fundamental challenge lies in systematically assessing LLMs' capabilities for specific CPS contexts. A pragmatic approach to this is engaging LLMs in structured dialogues, tailored to uncover their understanding and application of relevant concepts at different levels, including \textit{conceptual awareness} where the LLMs' fundamental conceptual grasp can be gauged by questions like "what is SSID in the context of WiFi?", and \textit{application and understanding} which delves deeper, examining whether LLMs can aptly apply fundamental concepts in practical scenarios with example questions like "what does it imply about the users' locations if their smartphones connect to WiFi APs with certain SSIDs and RSSIs?".



\textbf{Expanding LLMs' capabilities.} A subsequent and essential challenge is to broaden the capabilities of LLMs for CPS tasks based on the existing knowledge. Such expansion can be approached through several strategies.  \textit{Task decomposition} can break down complex tasks into simpler sub-tasks, which allows LLMs to develop more focused and efficient problem-solvers. \textit{Signal transformation and data preprocessing} decides the form in which sensor or actuator data shall be presented which is a crucial challenge. While digitized signals offer in-depth information, they require a deeper level of physical world understanding from LLMs. Transforming them into textualized data may be beneficial and other preprocessing methods such as filtering to remove irrelevant or redundant information may also enhance system efficacy. \textit{Effective prompt design} is a major challenge, which may involve embedding domain-specific knowledge when LLMs' common knowledge is limited in certain tasks. Developing stateful prompts and effective algorithms with fuzzy logic (as demonstrated in Section 3.2) is another interesting future work. \textit{Interfacing with external tools} also leads to an expansion of LLMs' capabilities. Examples include using code interpreters for executing signal processing algorithms or leveraging procedure calls for accessing real-time information and/or controlling CPS.


\textbf{Enriching LLMs with expert knowledge.} A pivotal approach is to develop specialized models tailored to embedding additional domain knowledge for CPS tasks. Such an approach however comes with special considerations and challenges: \textit{Dataset construction} for multimodal datasets to train tailored LLMs is a challenge. Unlike standard image-text pair datasets like those described in \cite{kakaobrain2022coyo}, sensor-text datasets for CPS tasks shall include not only descriptive information but also expert knowledge and processing guidance, which necessitates a thoughtful approach to ensure the data are comprehensive, accurate, and reflective of real-world scenarios. \textit{Balancing specialization with generalizability} is necessary. A critical risk in the fine-tuning LLMs is the potential disruption of the existing knowledge base of LLMs and a balanced fine-tuning process with both general and domain-specific data may be key to maintaining the robustness of LLMs. \textit{Integrating expert models} presents another way to enrich expert knowledge of LLMs, e.g., integrating LLMs with an IMU foundation model like LIMU-BERT \cite{xu2021limu} may enable frontend features of sensor signals before LLM comprehensions. 


\section{Related Work}
\textbf{LLMs for Sensor Processing.} With the scaling of model size and corpus size, LLMs demonstrate an emerging in-context learning (ICL) ability by learning directly from input prompts without additional training \citep{min2021metaicl,rubin2021learning,min2022rethinking}. This forward has broadened the application spectrum of LLMs, such as in Liu et al.'s study \citep{liu2023large}, where LLMs analyze medical data for health-related tasks, e.g., recognizing activities with accelerometer data. LLMs in \citep{liu2023large} primarily rely on learning from question-answer pairs presented in prompts. Our work, however, extends this by applying LLMs to signal-processing tasks, providing them with processed sensor data and structured guidance. We believe this enriched interaction between LLMs and sensor data can better exploit embedded common-sense knowledge in LLMs and thus unlock their potential to accomplish real-world tasks.



\textbf{Penetrative AI v.s. Embodied AI}. "Penetrative AI" is different from "Embodied AI" \citep{duan2022survey,ahn2022can,huang2022language,shridhar2020alfred,lin2023grounded}, which predominantly aims at designing robotic agents and is broadly defined with general AI models (rather than the penetrative AI's focus on LLMs' foundation roles). The penetrative AI focuses on the exploration of integrating LLMs with IoT sensing, which is not limited to the form of AI agents with tangible "bodies" and supports AGI-in-the-loop perception or control modules for CPS. While most Embodied AI works \citep{liang2023code, singh2023progprompt, lin2023grounded,wang2023voyager} are generally focused on executing everyday tasks (e.g., table cleaning), Penetrative AI is aimed at facilitating domain-specific tasks that require advanced expert knowledge, such as analyzing user behaviors. Many LLM-based embodied agents \citep{liang2023code, singh2023progprompt} center around the programming capabilities of LLMs, executing perceptions and actions through predefined API interfaces. In contrast to these robot-focused endeavors, this paper applies LLMs to directly comprehend physical world signals in IoT scenarios.

\textbf{Exploring Novel LLM Applications.} Several studies venture into novel LLM applications like image editing \citep{wu2023visual}, video understanding \citep{li2023videochat}, constructing knowledge graph \citep{sun2023think,carta2023iterative}, mental health prediction \citep{xu2023leveraging}, sequence completion \citep{generalpatternmachines2023}, and developing recommendation systems \citep{gao2023chat, liu2023chatgpt}. Different from all existing efforts, this paper defines "\textit{Penetrative AI}" which leverages LLMs' world knowledge in comprehending physical phenomena and completing real-world tasks. We believe this is the first effort to explore the boundaries of LLMs' ability to interact with the real physical world with IoT sensors.

\section{Conclusion}
We present penetrative AI and explore the potential of leveraging large language models as world models to accomplish real-world tasks with IoT sensors. Our findings illuminate a promising path for the integration of artificial intelligence and CPS, offering insights into the future of AI-powered solutions.

\section{Limitations}
Our study is based on a key assumption that LLMs have integrated high-level common-sense human knowledge that can be adopted for processing sensor data. This assumption may not be universally applicable to all LLMs, particularly those with small size or trained in specific NLP text corpora. Furthermore, our evaluation was confined to a select number of LLMs accessible through platforms such as OpenAI and Google API. This limited scope may not fully encompass the vast array of LLM capabilities currently available.

Due to constraints in manuscript length, we focused on two illustrative applications. While these were carefully chosen to represent distinct levels of signal processing within the Penetrative AI, they do not exhaust the full spectrum of potential applications. Despite this, we believe that these applications demonstrate the potential of LLMs in processing sensory signals. Future research could expand upon this groundwork by applying Penetrative AI to a wider array of applications.

We also observed that when employed in a penetrative manner, LLMs might exhibit lower efficiency in processing extensive sequences of digital data compared to traditional signal processing methods. This observation suggests a potential constraint in the practical deployment of Penetrative AI solutions. It underscores the need for continued research to enhance the efficiency of LLMs in handling long-digit sequences.

Since the three models assessed in this study are not open-source, it remains unclear if the data used for inference was included in their training datasets. However, we gather sensor data from our Android smartphones in the first application and ensure that all information remains confidential. In the second application, we utilize a publicly available dataset. Given that the raw signals undergo preprocessing (including random extraction, downsampling, and quantization), the possibility that the LLMs had previously encountered identical and long numerical sequences in their training datasets can be exceedingly low.

\section{Ethics Statement}
\textbf{Labor Considerations.} In constructing the dataset for activity sensing, authors and 7 volunteers engaged in tasks such as data collection and storage. Volunteers acknowledge the usage of data and collected sensor data are anonymous. The dataset includes human annotations that are fact-based, such as identifying whether the subject is indoors or outdoors during data collection. Thus, the sensor dataset maintains an objective and unbiased perspective.

\textbf{Misues Potential.} In our experiments with activity sensing, some LLMs demonstrated the capability to infer user activities from sensor data collected by smartphones. There exists a potential risk of future misuse, such as unauthorized tracking of users' location context information with satellite and WiFi signals as demonstrated in the first application. Although accessing those signals requires location permission from users in the higher versions of Android \citep{android@location, android@wifis}, we emphasize the necessity for responsible application of these technologies, with a strong commitment to protecting individual privacy and preventing malicious uses. Additionally, while applying LLMs such as ChatGPT-4 for heart rate detection holds promise, it necessitates further experimentation and studies to validate its effectiveness and reliability. Continued research in this area is crucial to ensure that LLMs can be confidently used for medical data analysis.

\section*{Acknowledgements}
We thank all reviewers for their insightful comments. This work is supported by Global STEM Professorship Scheme of Hong Kong and HKUST start up grant. It is also supported by Singapore NRF Investigatorship NRFI08-2022-0010. 

\bibliography{anthology,custom}

\begin{thebibliography}{51}
\expandafter\ifx\csname natexlab\endcsname\relax\def\natexlab#1{#1}\fi

\bibitem[{Ahn et~al.(2022)Ahn, Brohan, Brown, Chebotar, Cortes, David, Finn, Fu, Gopalakrishnan, Hausman et~al.}]{ahn2022can}
Michael Ahn, Anthony Brohan, Noah Brown, Yevgen Chebotar, Omar Cortes, Byron David, Chelsea Finn, Chuyuan Fu, Keerthana Gopalakrishnan, Karol Hausman, et~al. 2022.
\newblock Do as i can, not as i say: Grounding language in robotic affordances.
\newblock \emph{arXiv preprint arXiv:2204.01691}.

\bibitem[{Anil et~al.(2023)Anil, Dai, Firat, Johnson, Lepikhin, Passos, Shakeri, Taropa, Bailey, Chen et~al.}]{anil2023palm}
Rohan Anil, Andrew~M Dai, Orhan Firat, Melvin Johnson, Dmitry Lepikhin, Alexandre Passos, Siamak Shakeri, Emanuel Taropa, Paige Bailey, Zhifeng Chen, et~al. 2023.
\newblock Palm 2 technical report.
\newblock \emph{arXiv preprint arXiv:2305.10403}.

\bibitem[{Brown et~al.(2020)Brown, Mann, Ryder, Subbiah, Kaplan, Dhariwal, Neelakantan, Shyam, Sastry, Askell et~al.}]{brown2020language}
Tom Brown, Benjamin Mann, Nick Ryder, Melanie Subbiah, Jared~D Kaplan, Prafulla Dhariwal, Arvind Neelakantan, Pranav Shyam, Girish Sastry, Amanda Askell, et~al. 2020.
\newblock Language models are few-shot learners.
\newblock \emph{Advances in neural information processing systems}, 33.

\bibitem[{Byeon et~al.(2022)Byeon, Park, Kim, Lee, Baek, and Kim}]{kakaobrain2022coyo}
Minwoo Byeon, Beomhee Park, Haecheon Kim, Sungjun Lee, Woonhyuk Baek, and Saehoon Kim. 2022.
\newblock Coyo-700m: Image-text pair dataset.
\newblock \url{https://github.com/kakaobrain/coyo-dataset}.

\bibitem[{Carta et~al.(2023)Carta, Giuliani, Piano, Podda, Pompianu, and Tiddia}]{carta2023iterative}
Salvatore Carta, Alessandro Giuliani, Leonardo Piano, Alessandro~Sebastian Podda, Livio Pompianu, and Sandro~Gabriele Tiddia. 2023.
\newblock Iterative zero-shot llm prompting for knowledge graph construction.
\newblock \emph{arXiv preprint arXiv:2307.01128}.

\bibitem[{Christov(2004)}]{christov2004real}
Ivaylo~I Christov. 2004.
\newblock Real time electrocardiogram qrs detection using combined adaptive threshold.
\newblock \emph{Biomedical engineering online}, 3(1):1--9.

\bibitem[{Creswell et~al.(2022)Creswell, Shanahan, and Higgins}]{creswell2022selection}
Antonia Creswell, Murray Shanahan, and Irina Higgins. 2022.
\newblock Selection-inference: Exploiting large language models for interpretable logical reasoning.
\newblock \emph{arXiv preprint arXiv:2205.09712}.

\bibitem[{Developers(2023{\natexlab{a}})}]{android@gnss}
Android Developers. 2023{\natexlab{a}}.
\newblock \href {https://developer.android.com/reference/android/location/GnssStatus} {Gnssstatus}.

\bibitem[{Developers(2023{\natexlab{b}})}]{android@motion}
Android Developers. 2023{\natexlab{b}}.
\newblock \href {https://developer.android.com/develop/sensors-and-location/sensors/sensors_motion?hl=en#sensors-motion-stepcounter} {Motion sensors}.

\bibitem[{Developers(2023{\natexlab{c}})}]{android@location}
Android Developers. 2023{\natexlab{c}}.
\newblock \href {https://developer.android.com/develop/connectivity/wifi/wifi-scan} {Request location permission}.

\bibitem[{Developers(2023{\natexlab{d}})}]{android@wifi}
Android Developers. 2023{\natexlab{d}}.
\newblock \href {https://developer.android.com/reference/android/net/wifi/ScanResult} {Scanresult}.

\bibitem[{Developers(2023{\natexlab{e}})}]{android@wifis}
Android Developers. 2023{\natexlab{e}}.
\newblock \href {https://developer.android.com/develop/connectivity/wifi/wifi-scan} {Wifi scanning}.

\bibitem[{Duan et~al.(2022)Duan, Yu, Tan, Zhu, and Tan}]{duan2022survey}
Jiafei Duan, Samson Yu, Hui~Li Tan, Hongyuan Zhu, and Cheston Tan. 2022.
\newblock A survey of embodied ai: From simulators to research tasks.
\newblock \emph{IEEE Transactions on Emerging Topics in Computational Intelligence}, 6(2):230--244.

\bibitem[{Elgendi et~al.(2010)Elgendi, Jonkman, and De~Boer}]{elgendi2010frequency}
Mohamed Elgendi, Mirjam Jonkman, and Friso De~Boer. 2010.
\newblock Frequency bands effects on qrs detection.
\newblock \emph{Biosignals}, 2003:2002.

\bibitem[{Gao et~al.(2023)Gao, Sheng, Xiang, Xiong, Wang, and Zhang}]{gao2023chat}
Yunfan Gao, Tao Sheng, Youlin Xiang, Yun Xiong, Haofen Wang, and Jiawei Zhang. 2023.
\newblock Chat-rec: Towards interactive and explainable llms-augmented recommender system.
\newblock \emph{arXiv preprint arXiv:2303.14524}.

\bibitem[{Goldberger et~al.(2000)Goldberger, Amaral, Glass, Hausdorff, Ivanov, Mark, Mietus, Moody, Peng, and Stanley}]{goldberger2000physiobank}
Ary~L Goldberger, Luis~AN Amaral, Leon Glass, Jeffrey~M Hausdorff, Plamen~Ch Ivanov, Roger~G Mark, Joseph~E Mietus, George~B Moody, Chung-Kang Peng, and H~Eugene Stanley. 2000.
\newblock Physiobank, physiotoolkit, and physionet: components of a new research resource for complex physiologic signals.
\newblock \emph{circulation}, 101(23):e215--e220.

\bibitem[{Hamilton(2002)}]{hamilton2002open}
Pat Hamilton. 2002.
\newblock Open source ecg analysis.
\newblock In \emph{Computers in cardiology}, pages 101--104. IEEE.

\bibitem[{Huang et~al.(2022)Huang, Abbeel, Pathak, and Mordatch}]{huang2022language}
Wenlong Huang, Pieter Abbeel, Deepak Pathak, and Igor Mordatch. 2022.
\newblock Language models as zero-shot planners: Extracting actionable knowledge for embodied agents.
\newblock In \emph{International Conference on Machine Learning}, pages 9118--9147. PMLR.

\bibitem[{Jia et~al.(2021)Jia, Yang, Xia, Chen, Parekh, Pham, Le, Sung, Li, and Duerig}]{jia2021scaling}
Chao Jia, Yinfei Yang, Ye~Xia, Yi-Ting Chen, Zarana Parekh, Hieu Pham, Quoc Le, Yun-Hsuan Sung, Zhen Li, and Tom Duerig. 2021.
\newblock Scaling up visual and vision-language representation learning with noisy text supervision.
\newblock In \emph{International conference on machine learning}, pages 4904--4916. PMLR.

\bibitem[{Kadambe et~al.(1999)Kadambe, Murray, and Boudreaux-Bartels}]{kadambe1999wavelet}
Shubha Kadambe, Robin Murray, and G~Faye Boudreaux-Bartels. 1999.
\newblock Wavelet transform-based qrs complex detector.
\newblock \emph{IEEE Transactions on biomedical Engineering}, 46(7):838--848.

\bibitem[{Kalidas and Tamil(2017)}]{kalidas2017real}
Vignesh Kalidas and Lakshman Tamil. 2017.
\newblock Real-time qrs detector using stationary wavelet transform for automated ecg analysis.
\newblock In \emph{2017 IEEE 17th international conference on Bioinformatics and Bioengineering (BIBE)}, pages 457--461. IEEE.

\bibitem[{LeCun(2022)}]{lecun2022path}
Yann LeCun. 2022.
\newblock A path towards autonomous machine intelligence version 0.9. 2, 2022-06-27.
\newblock \emph{Open Review}, 62.

\bibitem[{Li et~al.(2023)Li, He, Wang, Li, Wang, Luo, Wang, Wang, and Qiao}]{li2023videochat}
KunChang Li, Yinan He, Yi~Wang, Yizhuo Li, Wenhai Wang, Ping Luo, Yali Wang, Limin Wang, and Yu~Qiao. 2023.
\newblock Videochat: Chat-centric video understanding.
\newblock \emph{arXiv preprint arXiv:2305.06355}.

\bibitem[{Liang et~al.(2023)Liang, Huang, Xia, Xu, Hausman, Ichter, Florence, and Zeng}]{liang2023code}
Jacky Liang, Wenlong Huang, Fei Xia, Peng Xu, Karol Hausman, Brian Ichter, Pete Florence, and Andy Zeng. 2023.
\newblock Code as policies: Language model programs for embodied control.
\newblock In \emph{2023 IEEE International Conference on Robotics and Automation (ICRA)}, pages 9493--9500. IEEE.

\bibitem[{Lin et~al.(2023)Lin, Huang, Liu, Gu, Sommerer, and Ren}]{lin2023grounded}
Bill~Yuchen Lin, Chengsong Huang, Qian Liu, Wenda Gu, Sam Sommerer, and Xiang Ren. 2023.
\newblock On grounded planning for embodied tasks with language models.
\newblock In \emph{Proceedings of the AAAI Conference on Artificial Intelligence}, volume~37, pages 13192--13200.

\bibitem[{Liu et~al.(2023{\natexlab{a}})Liu, Liu, Lv, Zhou, and Zhang}]{liu2023chatgpt}
Junling Liu, Chao Liu, Renjie Lv, Kang Zhou, and Yan Zhang. 2023{\natexlab{a}}.
\newblock Is chatgpt a good recommender? a preliminary study.
\newblock \emph{arXiv preprint arXiv:2304.10149}.

\bibitem[{Liu et~al.(2023{\natexlab{b}})Liu, McDuff, Kovacs, Galatzer-Levy, Sunshine, Zhan, Poh, Liao, Di~Achille, and Patel}]{liu2023large}
Xin Liu, Daniel McDuff, Geza Kovacs, Isaac Galatzer-Levy, Jacob Sunshine, Jiening Zhan, Ming-Zher Poh, Shun Liao, Paolo Di~Achille, and Shwetak Patel. 2023{\natexlab{b}}.
\newblock Large language models are few-shot health learners.
\newblock \emph{arXiv preprint arXiv:2305.15525}.

\bibitem[{Lu et~al.(2019)Lu, Batra, Parikh, and Lee}]{lu2019vilbert}
Jiasen Lu, Dhruv Batra, Devi Parikh, and Stefan Lee. 2019.
\newblock Vilbert: Pretraining task-agnostic visiolinguistic representations for vision-and-language tasks.
\newblock \emph{Advances in neural information processing systems}, 32.

\bibitem[{Min et~al.(2021)Min, Lewis, Zettlemoyer, and Hajishirzi}]{min2021metaicl}
Sewon Min, Mike Lewis, Luke Zettlemoyer, and Hannaneh Hajishirzi. 2021.
\newblock Metaicl: Learning to learn in context.
\newblock \emph{arXiv preprint arXiv:2110.15943}.

\bibitem[{Min et~al.(2022)}]{min2022rethinking}
Sewon Min et~al. 2022.
\newblock Rethinking the role of demonstrations: What makes in-context learning work?
\newblock \emph{arXiv preprint arXiv:2202.12837}.

\bibitem[{Mirchandani et~al.(2023)Mirchandani, Xia, Florence, Ichter, Driess, Arenas, Rao, Sadigh, and Zeng}]{generalpatternmachines2023}
Suvir Mirchandani, Fei Xia, Pete Florence, Brian Ichter, Danny Driess, Montserrat~Gonzalez Arenas, Kanishka Rao, Dorsa Sadigh, and Andy Zeng. 2023.
\newblock Large language models as general pattern machines.
\newblock In \emph{Proceedings of the 7th Conference on Robot Learning (CoRL)}.

\bibitem[{OpenAI(2023{\natexlab{a}})}]{gpt4_system_card}
OpenAI. 2023{\natexlab{a}}.
\newblock \href {https://openai.com/research/gpt-4v-system-card} {Gpt-4 system card}.

\bibitem[{OpenAI(2023{\natexlab{b}})}]{openai2023gpt4}
OpenAI. 2023{\natexlab{b}}.
\newblock \href {http://arxiv.org/abs/2303.08774} {Gpt-4 technical report}.

\bibitem[{Pan and Tompkins(1985)}]{pan1985real}
Jiapu Pan and Willis~J Tompkins. 1985.
\newblock A real-time qrs detection algorithm.
\newblock \emph{IEEE transactions on biomedical engineering}, (3):230--236.

\bibitem[{Popovi{\'c}(2015)}]{popovic2015chrf}
Maja Popovi{\'c}. 2015.
\newblock chrf: character n-gram f-score for automatic mt evaluation.
\newblock In \emph{Proceedings of the tenth workshop on statistical machine translation}, pages 392--395.

\bibitem[{Porr et~al.(2023)Porr, Howell, Stournaras, and Nir}]{bernd_porr_2023_7652725}
Bernd Porr, Luis Howell, Ioannis Stournaras, and Yoav Nir. 2023.
\newblock \href {https://doi.org/10.5281/zenodo.7652725} {Popular ecg r peak detectors written in python}.

\bibitem[{Radford et~al.(2021)Radford, Kim, Hallacy, Ramesh, Goh, Agarwal, Sastry, Askell, Mishkin, Clark et~al.}]{radford2021learning}
Alec Radford, Jong~Wook Kim, Chris Hallacy, Aditya Ramesh, Gabriel Goh, Sandhini Agarwal, Girish Sastry, Amanda Askell, Pamela Mishkin, Jack Clark, et~al. 2021.
\newblock Learning transferable visual models from natural language supervision.
\newblock In \emph{International conference on machine learning}, pages 8748--8763. PMLR.

\bibitem[{Rubin et~al.(2021)Rubin, Herzig, and Berant}]{rubin2021learning}
Ohad Rubin, Jonathan Herzig, and Jonathan Berant. 2021.
\newblock Learning to retrieve prompts for in-context learning.
\newblock \emph{arXiv preprint arXiv:2112.08633}.

\bibitem[{Scao et~al.(2022)Scao, Fan, Akiki, Pavlick, Ili{\'c}, Hesslow, Castagn{\'e}, Luccioni, Yvon, Gall{\'e} et~al.}]{scao2022bloom}
Teven~Le Scao, Angela Fan, Christopher Akiki, Ellie Pavlick, Suzana Ili{\'c}, Daniel Hesslow, Roman Castagn{\'e}, Alexandra~Sasha Luccioni, Fran{\c{c}}ois Yvon, Matthias Gall{\'e}, et~al. 2022.
\newblock Bloom: A 176b-parameter open-access multilingual language model.
\newblock \emph{arXiv preprint arXiv:2211.05100}.

\bibitem[{Shridhar et~al.(2020)Shridhar, Thomason, Gordon, Bisk, Han, Mottaghi, Zettlemoyer, and Fox}]{shridhar2020alfred}
Mohit Shridhar, Jesse Thomason, Daniel Gordon, Yonatan Bisk, Winson Han, Roozbeh Mottaghi, Luke Zettlemoyer, and Dieter Fox. 2020.
\newblock Alfred: A benchmark for interpreting grounded instructions for everyday tasks.
\newblock In \emph{Proceedings of the IEEE/CVF conference on computer vision and pattern recognition}, pages 10740--10749.

\bibitem[{Singh et~al.(2023)Singh, Blukis, Mousavian, Goyal, Xu, Tremblay, Fox, Thomason, and Garg}]{singh2023progprompt}
Ishika Singh, Valts Blukis, Arsalan Mousavian, Ankit Goyal, Danfei Xu, Jonathan Tremblay, Dieter Fox, Jesse Thomason, and Animesh Garg. 2023.
\newblock Progprompt: Generating situated robot task plans using large language models.
\newblock In \emph{2023 IEEE International Conference on Robotics and Automation (ICRA)}, pages 11523--11530. IEEE.

\bibitem[{Sun et~al.(2023)Sun, Xu, Tang, Wang, Lin, Gong, Shum, and Guo}]{sun2023think}
Jiashuo Sun, Chengjin Xu, Lumingyuan Tang, Saizhuo Wang, Chen Lin, Yeyun Gong, Heung-Yeung Shum, and Jian Guo. 2023.
\newblock Think-on-graph: Deep and responsible reasoning of large language model with knowledge graph.
\newblock \emph{arXiv preprint arXiv:2307.07697}.

\bibitem[{Tan and Bansal(2019)}]{tan2019lxmert}
Hao Tan and Mohit Bansal. 2019.
\newblock Lxmert: Learning cross-modality encoder representations from transformers.
\newblock \emph{arXiv preprint arXiv:1908.07490}.

\bibitem[{Wang et~al.(2023)Wang, Xie, Jiang, Mandlekar, Xiao, Zhu, Fan, and Anandkumar}]{wang2023voyager}
Guanzhi Wang, Yuqi Xie, Yunfan Jiang, Ajay Mandlekar, Chaowei Xiao, Yuke Zhu, Linxi Fan, and Anima Anandkumar. 2023.
\newblock Voyager: An open-ended embodied agent with large language models.
\newblock \emph{arXiv preprint arXiv:2305.16291}.

\bibitem[{Wei et~al.(2022)Wei, Wang, Schuurmans, Bosma, Xia, Chi, Le, Zhou et~al.}]{wei2022chain}
Jason Wei, Xuezhi Wang, Dale Schuurmans, Maarten Bosma, Fei Xia, Ed~Chi, Quoc~V Le, Denny Zhou, et~al. 2022.
\newblock Chain-of-thought prompting elicits reasoning in large language models.
\newblock \emph{Advances in Neural Information Processing Systems}, 35:24824--24837.

\bibitem[{Wu et~al.(2023)Wu, Yin, Qi, Wang, Tang, and Duan}]{wu2023visual}
Chenfei Wu, Shengming Yin, Weizhen Qi, Xiaodong Wang, Zecheng Tang, and Nan Duan. 2023.
\newblock Visual chatgpt: Talking, drawing and editing with visual foundation models.
\newblock \emph{arXiv preprint arXiv:2303.04671}.

\bibitem[{Xu et~al.(2021)Xu, Zhou, Tan, Li, and Shen}]{xu2021limu}
Huatao Xu, Pengfei Zhou, Rui Tan, Mo~Li, and Guobin Shen. 2021.
\newblock Limu-bert: Unleashing the potential of unlabeled data for imu sensing applications.
\newblock In \emph{Proceedings of the 19th ACM Conference on Embedded Networked Sensor Systems}, pages 220--233.

\bibitem[{Xu et~al.(2023)Xu, Yao, Dong, Yu, Hendler, Dey, and Wang}]{xu2023leveraging}
Xuhai Xu, Bingshen Yao, Yuanzhe Dong, Hong Yu, James Hendler, Anind~K Dey, and Dakuo Wang. 2023.
\newblock Leveraging large language models for mental health prediction via online text data.
\newblock \emph{arXiv preprint arXiv:2307.14385}.

\bibitem[{Yanowitz(2010)}]{yanowitz2010lesson}
Frank~G Yanowitz. 2010.
\newblock Lesson iii. characteristics of the normal ecg.
\newblock \emph{University of Utah School of Medicine}.

\bibitem[{Zeng et~al.(2022)Zeng, Liu, Du, Wang, Lai, Ding, Yang, Xu, Zheng, Xia et~al.}]{zeng2022glm}
Aohan Zeng, Xiao Liu, Zhengxiao Du, Zihan Wang, Hanyu Lai, Ming Ding, Zhuoyi Yang, Yifan Xu, Wendi Zheng, Xiao Xia, et~al. 2022.
\newblock Glm-130b: An open bilingual pre-trained model.
\newblock \emph{arXiv preprint arXiv:2210.02414}.

\bibitem[{Zhang et~al.(2019)Zhang, Kishore, Wu, Weinberger, and Artzi}]{zhang2019bertscore}
Tianyi Zhang, Varsha Kishore, Felix Wu, Kilian~Q Weinberger, and Yoav Artzi. 2019.
\newblock Bertscore: Evaluating text generation with bert.
\newblock In \emph{International Conference on Learning Representations}.

\end{thebibliography}

\appendix

\section{Complete Prompt}
\label{sec:app:pro}
\subsubsection{Activity Sensing}
In table~\ref{tab:exp:as}, we evaluate three prompt schemes for activity sensing: (1) plain prompt, (2) prompt with expert knowledge, and (3) prompt with expert knowledge and one reasoning example, which are shown in Figure \ref{fig:as:app:plain} to Figure \ref{fig:as:app:ek_ex}, respectively. All prompts include the objective, response format, and query sensor data. The response format is adopted to constrain the output of LLMs. We highlight placeholders for sensor data in blue and their detailed information is as follows:
\begin{itemize}
    \item \texttt{\$DATA\_STEP\$} represents the step count derived from the step counter algorithms.
    \item \texttt{\$DATA\_SATELLITE\_COUNT\$} indicates the satellite count.
    \item \texttt{\$DATA\_SATELLITE\_SNR\$} is the average SNR of satellite signals.
    \item \texttt{\$DATA\_WIFI\_COUNT\$} denotes the count of WiFi APs with an RSSI above -70.
    \item \texttt{\$DATA\_WIFI\_LIST\$} indicates the SSID list of WiFi APs with RSSI over -70.
\end{itemize}
All placeholders are replaced with actual sensor data for new inference. For instance, "\texttt{\$DATA\_STEP\$}" might be replaced by "5.2", resulting in the complete phrase "Step count: 5.2/min." 

\subsubsection{Heart Rate Detection}
Figure~\ref{fig:hd:app:desc} to Figure~\ref{fig:hd:app:proc_ex} present the prompt templates to LLMs for the R-peak detection task. Similarly, each prompt template incorporates a response format and a placeholder for ECG digits. Notably, a special sentence - "Do not write codes" is inserted in the prompt to prevent LMs from generating code as a solution. Figure~\ref{fig:hd:app:data} demonstrates an example of query ECG data, which can be used to replace the placeholder in the prompt templates and get the complete prompt.

\begin{figure*}
\begin{subfigure}{1.0\linewidth}
  \centering
  \includegraphics[width=\linewidth]{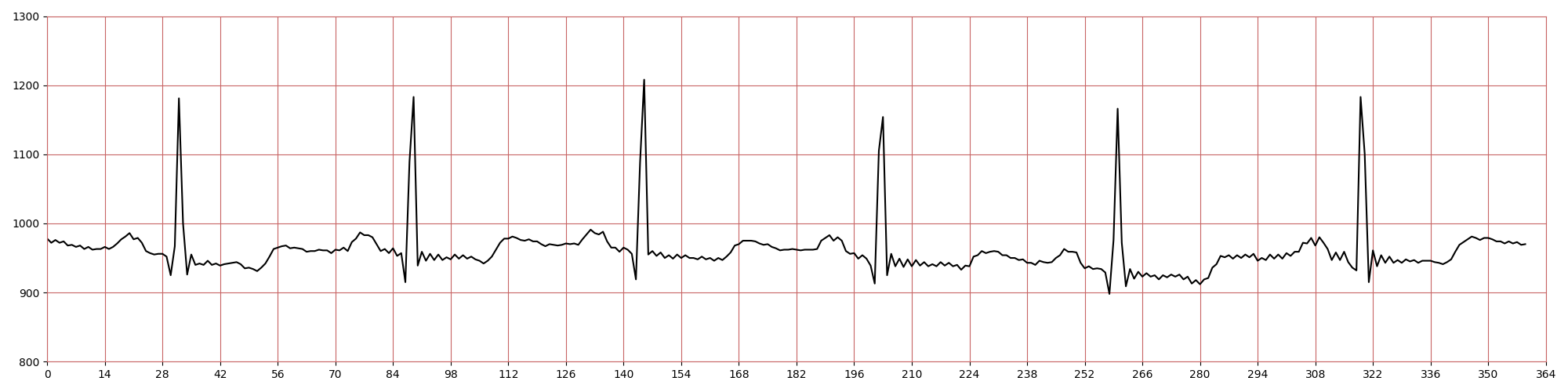}
  \caption{Reference figure for reasoning example.}
\end{subfigure}
\\
\begin{subfigure}{1.00\linewidth}
  \centering
  \includegraphics[width=\linewidth]{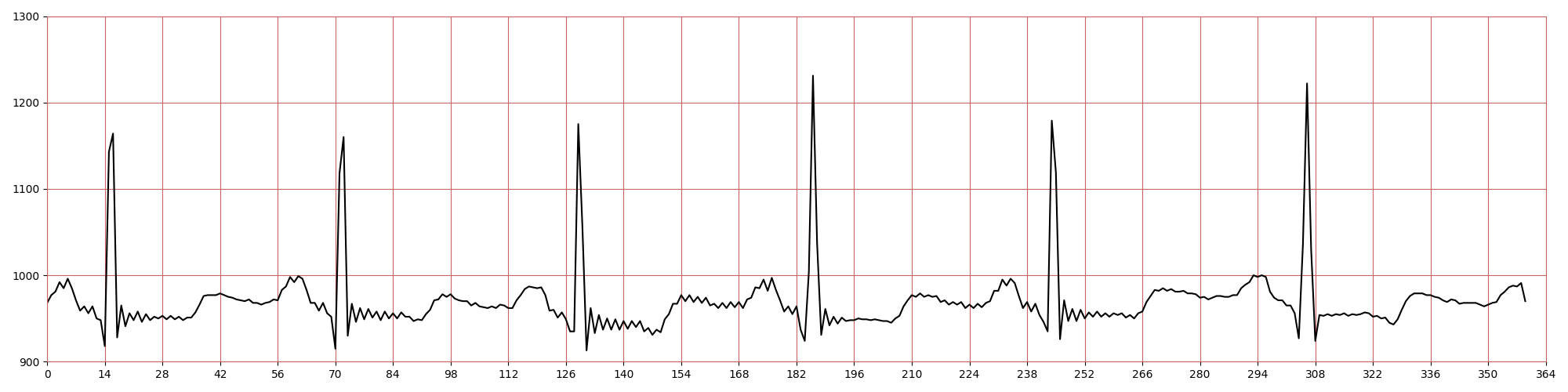}
  \caption{Query figure.}
\end{subfigure}
\caption{ECG Figure examples for VLMs.}
\label{fig:hdv:img}
\end{figure*}

Figure~\ref{fig:hdv:app:des} presents the prompt template with descriptions of R-peak for heart rate detection with images, while Figure~\ref{fig:hdv:app:proc_ex} displays prompt templates incorporating a description or a reasoning example. Since the digited ECG data are input as figures, we omit the textual placeholder. In practice, we may input multiple figures into VLMs, including a query figure and reference images for reasoning examples.  Figure~\ref{fig:hdv:img} showcases examples of ECG data figures, formatted in PNG and sized at $2000\times500$.

\begin{figure*}[!htb]
\begin{mdframed}[backgroundcolor=gray!20, linewidth=1pt, linecolor=black, innertopmargin=\baselineskip, innerbottommargin=\baselineskip]
\footnotesize

Objective:\\
Determine a user's activity by analyzing sensor data from their smartphone.\\ 

Response Format:\\
Reasoning: Provide a comprehensive analysis of the sensor data. \\
Summary: Conclude with a brief summary of your findings. \\
Motion: choose one from either 'stationary' or 'walking'. \\
Environment: choose one from either 'indoors' or 'outdoors'.\\ 

Now infer a user's motion and surrounding conditions with the following sensor data:\\
Sensor data:\\
1. Step count: \textcolor{blue}{\$DATA_STEP\$}/min.\\
2. Satellites detected: \textcolor{blue}{\$DATA_SATELLITE_COUNT\$}. Carrier-to-noise: \textcolor{blue}{\$DATA_SATELLITE_SNR\$}dB.\\
3. Total WiFi APs scanned: \textcolor{blue}{\$DATA_WIFI_COUNT\$}. SSID list: \textcolor{blue}{\$DATA_WIFI_LIST\$}.\\
Reasoning:\\
Summary:\\
Motion:\\
Environment:
\end{mdframed}
\captionof{figure}{Prompt template (plain) for activity sensing.}
\label{fig:as:app:plain}
\end{figure*}

\begin{figure*}[!htb]
\begin{mdframed}[backgroundcolor=gray!20, linewidth=1pt, linecolor=black, innertopmargin=\baselineskip, innerbottommargin=\baselineskip]
\footnotesize

Objective:\\
Determine a user's activity by analyzing sensor data from their smartphone.\\ 

Sensor Data and Expert Knowledge:\\
You will receive data from various sensors, including the accelerometer, satellite, and WiFi. Here's how to interpret this data:\\
1. Step Count per Minute:\\
Source: Accelerometer (measures user's movement).\\
Interpretation: A high count signifies walking; a low count indicates the user is likely stationary.\\
2. Satellite Data:\\
Data: Number of satellites detected and average carrier-to-noise density (in dB).\\
Interpretation: High satellite count and carrier-to-noise density indicates an outdoor setting with strong satellite signals.\\
3. WiFi Data:\\
Data: Total count of WiFi Access Points (APs) detected and the list of their SSID.\\
Interpretation:
A large total count of detected APs implies that the user is likely in close proximity to or inside a building, given the prevalence of WiFi in modern buildings.
Scanned APs indicate user's proximity to them, and their SSIDs can hint at specific locations. So analyze each SSID. For example, an SSID named 'Starbucks' suggests the user is close to a Starbucks. Note: Some SSIDs may be not meaningful.\\

Response Format:\\
Reasoning: Provide a comprehensive analysis of the sensor data. \\
Summary: Conclude with a brief summary of your findings. \\
Motion: choose one from either 'stationary' or 'walking'. \\
Environment: choose one from either 'indoors' or 'outdoors'.\\ 

Now infer a user's motion and surrounding conditions with the following sensor data:\\
Sensor data:\\
1. Step count: \textcolor{blue}{\$DATA_STEP\$}/min.\\
2. Satellites detected: \textcolor{blue}{\$DATA_SATELLITE_COUNT\$}. Carrier-to-noise: \textcolor{blue}{\$DATA_SATELLITE_SNR\$}dB.\\
3. Total WiFi APs scanned: \textcolor{blue}{\$DATA_WIFI_COUNT\$}. SSID list: \textcolor{blue}{\$DATA_WIFI_LIST\$}.\\
Reasoning:\\
Summary:\\
Motion:\\
Environment:
\end{mdframed}
\captionof{figure}{Prompt template (with expert knowledge) for activity sensing.}
\label{fig:as:app:ek}
\end{figure*}

\begin{figure*}[!htb]
\begin{mdframed}[backgroundcolor=gray!20, linewidth=1pt, linecolor=black, innertopmargin=\baselineskip, innerbottommargin=\baselineskip]
\footnotesize

Objective:\\
Determine a user's activity by analyzing sensor data from their smartphone.\\ 

Sensor Data and Expert Knowledge:\\
You will receive data from various sensors, including the accelerometer, satellite, and WiFi. Here's how to interpret this data:\\
1. Step Count per Minute:\\
Source: Accelerometer (measures user's movement).\\
Interpretation: A high count signifies walking; a low count indicates the user is likely stationary.\\
2. Satellite Data:\\
Data: Number of satellites detected and average carrier-to-noise density (in dB).\\
Interpretation: High satellite count and carrier-to-noise density indicates an outdoor setting with strong satellite signals.\\
3. WiFi Data:\\
Data: Total count of WiFi Access Points (APs) detected and the list of their SSID.\\
Interpretation:
A large total count of detected APs implies that the user is likely in close proximity to or inside a building, given the prevalence of WiFi in modern buildings.
Scanned APs indicate user's proximity to them, and their SSIDs can hint at specific locations. So analyze each SSID. For example, an SSID named 'Starbucks' suggests the user is close to a Starbucks. Note: Some SSIDs may be not meaningful.\\

Response Format:\\
Reasoning: Provide a comprehensive analysis of the sensor data. \\
Summary: Conclude with a brief summary of your findings. \\
Motion: choose one from either 'stationary' or 'walking'. \\
Environment: choose one from either 'indoors' or 'outdoors'.\\ 

Reasoning Example:\\
Sensor Data:\\
1. Step count: 5/min.\\
2. Satellites detected: 16. Carrier-to-noise: 35.46dB.\\
3. Total WiFi APs scanned: 6. SSID list: ['McDonald's Singapore', 'xiaomi_5G', 'McDonald's Singapore', 'Android_xx123', 'OPPO 196', 'link-B33'].\\
Reasoning:\\
1. The low step count indicates the user is stationary.\\
2. A high number of detected satellites and high carrier-to-noise suggest an outdoor environment.\\
3. WiFi data shows some WiFi Access Points (APs) detected, with SSIDs like 'McDonald's Singapore' hinting at a location close to McDonald restaurant in Singapore.\\
Summary: The user is stationary, likely in an outdoor area near a McDonald restaurant in Singapore.\\
Motion: stationary.\\
Environment: indoors.\\

Now infer a user's motion and surrounding conditions with the following sensor data:\\
Sensor data:\\
1. Step count: \textcolor{blue}{\$DATA_STEP\$}/min.\\
2. Satellites detected: \textcolor{blue}{\$DATA_SATELLITE_COUNT\$}. Carrier-to-noise: \textcolor{blue}{\$DATA_SATELLITE_SNR\$}dB.\\
3. Total WiFi APs scanned: \textcolor{blue}{\$DATA_WIFI_COUNT\$}. SSID list: \textcolor{blue}{\$DATA_WIFI_LIST\$}.\\
Reasoning:\\
Summary:\\
Motion:\\
Environment:
\end{mdframed}
\captionof{figure}{Prompt template (with expert knowledge) for activity sensing.}
\label{fig:as:app:ek_ex}
\end{figure*}

\begin{figure*}[!htb]
\begin{mdframed}[backgroundcolor=gray!20, linewidth=1pt, linecolor=black, innertopmargin=\baselineskip, innerbottommargin=\baselineskip]
\footnotesize

Objective:\\
Find the R-peaks in an ECG waveform.\\

Background Knowledge:\\
The QRS complex, a recurring feature in ECG data, signifies the ventricles' consistent depolarization in the heart. It comprises the Q, R, and S waves, where the Q wave shows a downward deflection, followed by an upward-moving R wave, and then the S wave, which deflects downward after the R wave. The maximum amplitude of the R wave is known as the R-peak.\\

Response Format:\\
Your response should strictly adhere to the format detailed below:\\
Reasoning: Provide a reasoned explanation based on the information mentioned above about how the R-peaks were identified.\\
R-peaks: List the identified R-peak values in the format [R1, R2, R3], including duplicates as separate entries.\\

Please identify the R-peaks in the provided ECG data. Do not write codes.\\
ECG data: \textcolor{blue}{\$DATA\$}
\end{mdframed}
\captionof{figure}{Prompt template (with descriptions) for R-peak detection.}
\label{fig:hd:app:desc}
\end{figure*}

\begin{figure*}[!htb]
\begin{mdframed}[backgroundcolor=gray!20, linewidth=1pt, linecolor=black, innertopmargin=\baselineskip, innerbottommargin=\baselineskip]
\footnotesize

Objective:\\
Find the R-peaks in an ECG waveform.\\

Background Knowledge:\\
An R-peak within a sequence of ECG numbers refers to a pronounced upward deflection, typically representing the largest and most conspicuous values within the sequence. To identify R-peaks, follow these steps:\\

1. Initial Observation: Begin by observing the rough overall range of ECG numbers in the provided data.\\

2. Identify Subsequences: Find subsequences of numbers that meet the following criteria:\\
2.1. The initial numbers are in the lower part of the overall range, even smaller than the range.\\
2.2. Subsequent numbers exhibit a significant increase, even exceeding the overall range.\\
2.3. Following the increase, subsequent numbers quickly return to the lower part of the overall range.\\

3. Select R-Peaks: After identifying these subsequences, select the largest number from each subsequence as an R-peak.\\

Response Format:\\
Your response should strictly adhere to the format detailed below:\\
Reasoning: Provide a reasoned explanation based on the information mentioned above about how the R-peaks were identified.\\
R-peaks: List the identified R-peak values in the format [R1, R2, R3], including duplicates as separate entries.\\

Please identify the R-peaks in the provided ECG data. Do not write codes.\\
ECG data: \textcolor{blue}{\$DATA\$}
\end{mdframed}
\captionof{figure}{Prompt template (with a procedure) for R-peak detection.}
\label{fig:hd:app:proc}
\end{figure*}

\begin{figure*}[!htb]
\begin{mdframed}[backgroundcolor=gray!20, linewidth=1pt, linecolor=black, innertopmargin=\baselineskip, innerbottommargin=\baselineskip]
\footnotesize

Objective:\\
Find the R-peaks in an ECG waveform.\\

Background Knowledge:\\
An R-peak within a sequence of ECG numbers refers to a pronounced upward deflection, typically representing the largest and most conspicuous values within the sequence. To identify R-peaks, follow these steps:\\

1. Initial Observation: Begin by observing the rough overall range of ECG numbers in the provided data.\\

2. Identify Subsequences: Find subsequences of numbers that meet the following criteria:\\
2.1. The initial numbers are in the lower part of the overall range, even smaller than the range.\\
2.2. Subsequent numbers exhibit a significant increase, even exceeding the overall range.\\
2.3. Following the increase, subsequent numbers quickly return to the lower part of the overall range.\\

3. Select R-Peaks: After identifying these subsequences, select the largest number from each subsequence as an R-peak.\\

Response Format:\\
Your response should strictly adhere to the format detailed below:\\
Reasoning: Provide a reasoned explanation based on the information mentioned above about how the R-peaks were identified.\\
R-peaks: List the identified R-peak values in the format [R1, R2, R3], including duplicates as separate entries.\\

Reasoning Example:\\
ECG data: [978, 972, 976, 972, 974, 968, 969, 966, 968, 963, 966, 962, 963, 963, 966, 963, 966, 971, 977, 981, 986, 977, 979, 972, 960, 957, 955, 956, 956, 952, 925, 967, 1181, 1000, 926, 955, 940, 942, 940, 946, 940, 942, 939, 941, 942, 943, 944, 941, 935, 936, 934, 931, 936, 942, 952, 963, 965, 967, 968, 964, 965, 964, 963, 959, 960, 960, 962, 961, 961, 957, 962, 961, 965, 960, 973, 978, 987, 983, 983, 980, 970, 960, 963, 957, 964, 953, 957, 915, 1089, 1183, 939, 959, 946, 956, 947, 955, 947, 951, 948, 955, 949, 954, 949, 952, 948, 946, 942, 946, 952, 962, 972, 978, 978, 981, 979, 976, 975, 977, 974, 974, 970, 967, 970, 969, 968, 969, 971, 970, 971, 969, 977, 984, 991, 986, 984, 988, 974, 965, 965, 959, 965, 962, 956, 919, 1088, 1208, 955, 960, 953, 958, 950, 954, 949, 955, 950, 954, 950, 950, 948, 952, 948, 950, 946, 950, 947, 952, 958, 968, 970, 975, 975, 975, 974, 971, 969, 970, 966, 964, 961, 962, 962, 963, 962, 961, 962, 962, 962, 963, 975, 979, 983, 975, 980, 975, 960, 956, 957, 949, 954, 949, 939, 913, 1105, 1154, 925, 956, 938, 949, 937, 948, 938, 947, 939, 944, 938, 941, 938, 944, 939, 943, 938, 940, 933, 939, 938, 952, 954, 960, 957, 959, 960, 959, 954, 954, 950, 950, 947, 948, 943, 943, 940, 946, 944, 943, 944, 950, 954, 963, 959, 959, 958, 943, 935, 938, 934, 935, 934, 929, 898, 976, 1166, 972, 909, 934, 920, 930, 923, 928, 923, 925, 919, 925, 922, 926, 923, 926, 919, 923, 913, 918, 912, 919, 921, 936, 941, 953, 951, 954, 949, 954, 950, 955, 951, 956, 946, 950, 947, 955, 949, 955, 949, 957, 953, 959, 959, 972, 971, 979, 968, 980, 972, 963, 947, 958, 947, 959, 944, 936, 932, 1183, 1101, 915, 961, 938, 954, 943, 952, 943, 947, 943, 948, 945, 947, 943, 946, 946, 946, 944, 943, 941, 944, 948, 959, 969, 973, 977, 981, 979, 976, 979, 979, 977, 974, 974, 971, 974, 971, 973, 969, 970]\\
Reasoning:\\
Following the three steps by:\\
1. Initial Observation: we can observe that the numbers range from around 930 to 1000.\\
2. Identify Subsequences: we can identify several subsequences that start with numbers in the lower part of the range, significantly increase to exceed the range, and then return to the range:\\
- [925, 967, 1181, 1000, 926]\\
- [915, 1089, 1183, 939]\\
- [919, 1088, 1208, 955]\\
- [913, 1105, 1154, 925]\\
- [898, 976, 1166, 972, 909]\\
- [932, 1183, 1101, 915]\\
3. Select R-Peaks: The largest number from those subsequences are [1181, 1183, 1208, 1154, 1166, 1183].\\
R-peaks: [1181, 1183, 1208, 1154, 1166, 1183].\\

Please identify the R-peaks in the provided ECG data. Do not write codes.\\
ECG data: \textcolor{blue}{\$DATA\$}
\end{mdframed}
\captionof{figure}{Prompt template (with a procedure and a reasoning example) for R-peak detection.}
\label{fig:hd:app:proc_ex}
\end{figure*}

\begin{figure*}[!htb]
\begin{mdframed}[backgroundcolor=green!20, linewidth=1pt, linecolor=black, innertopmargin=\baselineskip, innerbottommargin=\baselineskip]
\footnotesize

[968, 977, 981, 992, 985, 996, 985, 971, 959, 964, 956, 964, 950, 948, 918, 1143, 1164, 928, 965, 941, 956, 948, 958, 946, 955, 948, 952, 950, 953, 949, 953, 949, 952, 948, 951, 951, 957, 966, 976, 977, 977, 977, 979, 977, 975, 974, 972, 971, 970, 972, 968, 968, 966, 968, 969, 972, 971, 983, 987, 998, 992, 999, 996, 983, 968, 968, 959, 968, 956, 952, 915, 1118, 1160, 930, 967, 946, 962, 949, 961, 951, 958, 948, 958, 950, 956, 950, 957, 952, 952, 947, 949, 948, 955, 960, 971, 972, 978, 975, 978, 973, 971, 970, 970, 965, 968, 964, 963, 962, 964, 962, 966, 965, 962, 962, 971, 977, 984, 987, 986, 985, 986, 977, 959, 960, 951, 957, 949, 935, 935, 1175, 1056, 913, 962, 933, 954, 937, 950, 937, 949, 937, 947, 938, 947, 940, 947, 935, 939, 931, 937, 934, 949, 955, 967, 967, 977, 970, 977, 969, 975, 968, 974, 965, 967, 962, 968, 962, 969, 963, 969, 962, 972, 974, 986, 985, 995, 982, 997, 983, 971, 958, 964, 955, 964, 937, 924, 1004, 1231, 1037, 931, 961, 942, 952, 944, 951, 947, 948, 948, 950, 949, 949, 948, 949, 948, 947, 947, 945, 950, 953, 964, 971, 977, 975, 979, 975, 977, 975, 976, 969, 971, 966, 969, 966, 969, 962, 966, 962, 967, 963, 968, 970, 982, 982, 995, 988, 996, 991, 976, 962, 969, 958, 967, 954, 946, 935, 1179, 1119, 926, 971, 947, 961, 947, 960, 950, 957, 952, 958, 952, 956, 952, 956, 954, 956, 951, 954, 950, 956, 958, 969, 976, 983, 982, 985, 982, 984, 981, 981, 982, 979, 979, 978, 974, 975, 972, 974, 976, 976, 975, 975, 977, 977, 985, 989, 992, 1000, 998, 1000, 998, 981, 974, 971, 971, 965, 965, 956, 927, 1036, 1222, 1029, 924, 954, 953, 955, 953, 955, 954, 956, 953, 955, 954, 955, 957, 956, 952, 953, 950, 951, 945, 943, 949, 960, 970, 976, 979, 979, 979, 977, 977, 975, 974, 971, 969, 972, 971, 967, 968, 968, 968, 968, 966, 964, 966, 968, 969, 977, 981, 986, 988, 987, 991, 970]

\end{mdframed}
\captionof{figure}{Example query ECG data for R-peak detection.}
\label{fig:hd:app:data}
\end{figure*}

\begin{figure*}[!htb]
\begin{mdframed}[backgroundcolor=gray!20, linewidth=1pt, linecolor=black, innertopmargin=\baselineskip, innerbottommargin=\baselineskip]
\footnotesize

Objective: \\
Find R-peaks in an ECG waveform.\\

Background Knowledge:\\
The QRS complex, a recurring feature in ECG data, signifies the ventricles' consistent depolarization in the heart. It comprises the Q, R, and S waves, where the Q wave shows a downward deflection, followed by an upward-moving R wave, and then the S wave, which deflects downward after the R wave. The maximum amplitude of the R wave is known as the R-peak.

Response Format:\\
Your response should strictly adhere to the format detailed below:\\
Reasoning: Provide a reasoned explanation based on the information mentioned above about how the R-peaks were identified.\\
R-peaks: List the approximate indices of identified R-peak in the format [R1, R2, R3], including duplicates as separate entries.\\

You should utilize the procedures described above, count R-peaks in the second image. In this task, the use of coding for identification or counting of R-peaks is not permitted.
\end{mdframed}
\captionof{figure}{Vison prompt template (with a procedure and a reasoning example) for R-peak detection.}
\label{fig:hdv:app:des}
\end{figure*}

\begin{figure*}[!htb]
\begin{mdframed}[backgroundcolor=gray!20, linewidth=1pt, linecolor=black, innertopmargin=\baselineskip, innerbottommargin=\baselineskip]
\footnotesize

Objective: \\
Find R-peaks in an ECG waveform.\\

Background Knowledge:\\
An R-peak within a sequence of ECG numbers refers to a pronounced upward deflection, typically representing the largest and most conspicuous values within the sequence. To identify R-peaks, follow these procedures:\\
1.Identification of R-wave Spikes: Initially, identify the distinct sharp spikes in the ECG waveform that signify the R waves.\\
2.Vertex Determination and Coordinate Extraction: Subsequently, determine the vertices of these identified spikes. Then Extract the x-axis coordinates of these vertices.\\

Response Format:\\
Your response should strictly adhere to the format detailed below:\\
Reasoning: Provide a reasoned explanation based on the information mentioned above about how the R-peaks were identified.\\
R-peaks: List the approximate indices of identified R-peak in the format [R1, R2, R3], including duplicates as separate entries.\\

Reasoning Example:\\
For the first provided image, following the steps by:\\
1. There are six sharp spikes in total.\\
2. The coordinates are approximately [32,89,145,203,260,320].\\

You should utilize the procedures described above, count R-peaks in the second image. In this task, the use of coding for identification or counting of R-peaks is not permitted.
\end{mdframed}
\captionof{figure}{Vison prompt template (with a procedure and a reasoning example) for R-peak detection.}
\label{fig:hdv:app:proc_ex}
\end{figure*}

\end{document}